\documentclass[10pt,twocolumn,letterpaper]{article}

\usepackage[pagenumbers]{cvpr} 

\definecolor{cvprblue}{rgb}{0.21,0.49,0.74}
\usepackage[pagebackref,breaklinks,colorlinks,allcolors=cvprblue]{hyperref}
\usepackage{bm}
\usepackage{color}
\usepackage{cuted}
\usepackage{etoolbox}
\usepackage{colortbl}
\usepackage{graphicx}
\usepackage{makecell}
\usepackage{threeparttable}
\usepackage{booktabs}
\usepackage{amsmath, amssymb}
\usepackage{nicefrac}
\usepackage[accsupp]{axessibility}  

\def\paperID{2846} 
\def\confName{CVPR}
\def\confYear{2025}

\title{Ref-GS: Directional Factorization for 2D Gaussian Splatting}

\author{
    Youjia Zhang$^{1}$\quad
    Anpei Chen$^{2,3, \dag}$ \quad
    Yumin Wan$^{1}$ \quad
    Zikai Song$^{1}$ \quad \\
    Junqing Yu$^{1}$ \quad
    Yawei Luo$^{4}$ \quad
    Wei Yang$^{1, \dag}$
    \vspace{0.2cm}
\\
    {\normalsize $^{1}$ Huazhong University of Science and Technology} \quad
    {\normalsize $^{2}$ University of Tübingen, Tübingen AI Center} \quad \\
    {\normalsize $^{3}$ Westlake University} \quad 
    {\normalsize $^{4}$ Zhejiang University}
}

\begin{document}

\maketitle

\begin{strip}
    \vspace{-1.2cm}
    \centering
    \includegraphics[width=\textwidth]{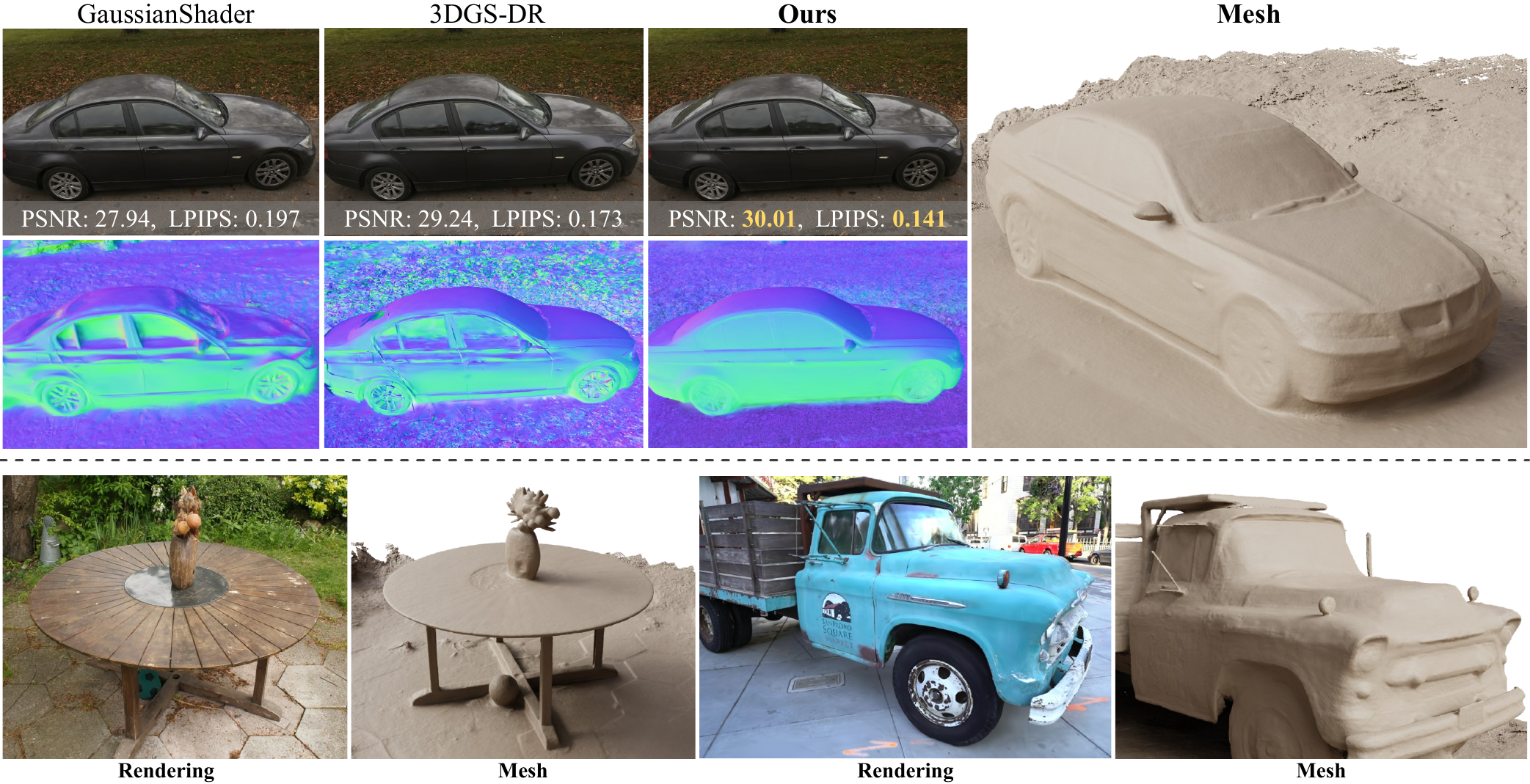}
    \captionof{figure}{
    Our \textit{Ref-GS} method generates photo-realistic renderings with view-dependent effects while also enabling accurate geometry recovery. The top row shows a comparison of renderings for a scene with specular reflections, along with the recovered normals and mesh. The bottom row demonstrates our successful reconstruction of the geometries of the `reflective table center' in the \textit{`Garden'} scene~\cite{barron2022mipnerf360} and the `windshield' in the \textit{`Truck'} scene~\cite{knapitsch2017tanks}, which existing methods typically fail to handle.
    }
    \label{fig:teaser}
\end{strip}

\renewcommand{\thefootnote}{\fnsymbol{footnote}}
\footnotetext{$\dag$ denote co-corresponding authors.}

\begin{abstract}
In this paper, we introduce \textit{Ref-GS}, a novel approach for directional light factorization in 2D Gaussian splatting~\cite{Huang2DGS2024}, which enables photorealistic view-dependent appearance rendering and precise geometry recovery. %
\textit{Ref-GS} builds upon the deferred rendering of Gaussian splatting and applies directional encoding to the deferred-rendered surface, effectively reducing the ambiguity between orientation and viewing angle. Next, we introduce a spherical Mip-grid to capture varying levels of surface roughness, enabling roughness-aware Gaussian shading. %
Additionally, we propose a simple yet efficient geometry-lighting factorization that connects geometry and lighting via the vector outer product, significantly reducing renderer overhead when integrating volumetric attributes. %
Our method achieves superior photorealistic rendering for a range of open-world scenes while also accurately recovering geometry. See our interactive {\href{https://ref-gs.github.io/}{\textbf{project page}.}}
\end{abstract}    
\section{Introduction}

\label{sec:intro}
View-dependent effects are a key element in 3D reconstruction and rendering - capturing complex interactions of light in materials such as reflection and refraction - have been studied for decades in forward rendering for computer graphics to enhance realism and visual fidelity in simulations, animations, and visual effects.
Recent advances in Neural Radiance Fields (NeRF~\cite{mildenhall2020nerf}, Mildenhall 
\etal in 2020) and Gaussian Splatting (GS~\cite{kerbl3Dgaussians}, Kerbl \etal in 2023) have enabled high-fidelity 3D scene reconstruction and novel view synthesis. However, unlike forward rendering in computer graphics, light propagation such as reflection and refraction has received much less attention in neural fields, and significant artifacts in both geometry and rendered images can be observed when reconstructing scenes with complex materials.

This is because NeRF and its follow-ups~\cite {barron2021mipnerf, barron2022mipnerf360, wang2021neus, yariv2021volume} represent 3D scenes as a collection of emission radiance points and query view-dependent colors using viewing direction, without accounting for the bouncing and bending of light rays as they travel from the light source to the viewing cameras. To tackle this issue, Ref-NeRF~\cite{verbin2022refnerf} take advantage of surface light field rendering~\cite{slf, dslf} and replaces NeRF's directional parameterization with an integrated reflection encoding, achieving significant improvement in the realism and accuracy of specular reflections. Recent work~\cite{wu2024neural} brings feature grid-based encoding to the directional domain to speed up the efficiency of directional encoding. In the context of 3DGS, directly applying the reflection of the view direction as the view-dependent color query for recent efficient GS representation is problematic, as it independently inherits model orientation and Spherical Harmonics (SH) color for each primitive, resulting in transforming viewing direction can easily be offset during parameter updating. To do so, recent work~\cite{yang2024spec,ye2024gsdr,wang2024specgaussian} incorporates smooth regularization and higher-order view-dependent color modeling into the rendering function, achieving promising quality on reflective surfaces. Despite the high rendering quality, they are struggling to provide accurate geometry.

In this paper, we present \textit{Ref-GS}, a new directional encoding method for 2D Gaussian splatting that leverages deferred rendering and lighting factorization to achieve photorealistic view-dependent effect reconstruction, while preserving accurate geometry. Unlike previous work that treats ray color as the integration of point radiance, we leverage deferred rendering techniques by postponing view-dependent color evaluation until after Gaussian attribute blending and performing directional encoding only on the estimated surface, which efficiently reduces the orientation-viewing ambiguity of Gaussian representations (see Section~\ref{sec:ambiguty}).
In addition, we introduce a Mip-grid to capture varying levels of surface roughness, enabling roughness-aware Gaussian shading.
Furthermore, spatially varying materials are crucial for modeling open-world scenes; therefore, we propose a simple yet efficient geometry-lighting factorization that connects the geometry and lighting through the vector outer product, significantly lowering renderer overhead for volumetric attribute integration.

Our \textit{Ref-GS} achieves effective reconstruction of high-frequency reflection and fraction from multi-view images and enables faithful geometry recovery. An extensive evaluation of our approach with both synthetic and real-world scenes demonstrates that \textit{Ref-GS} produces state-of-the-art renderings of novel views, even compared to implicit methods. Furthermore, our method maintains competitive training times and importantly allows high-quality real-time ($>$ 45 FPS) novel view synthesis at $800 \times 800$ resolution with a novel deferred mechanism.

\begin{figure*}[t]
\centering
\includegraphics[width=1.0\linewidth]{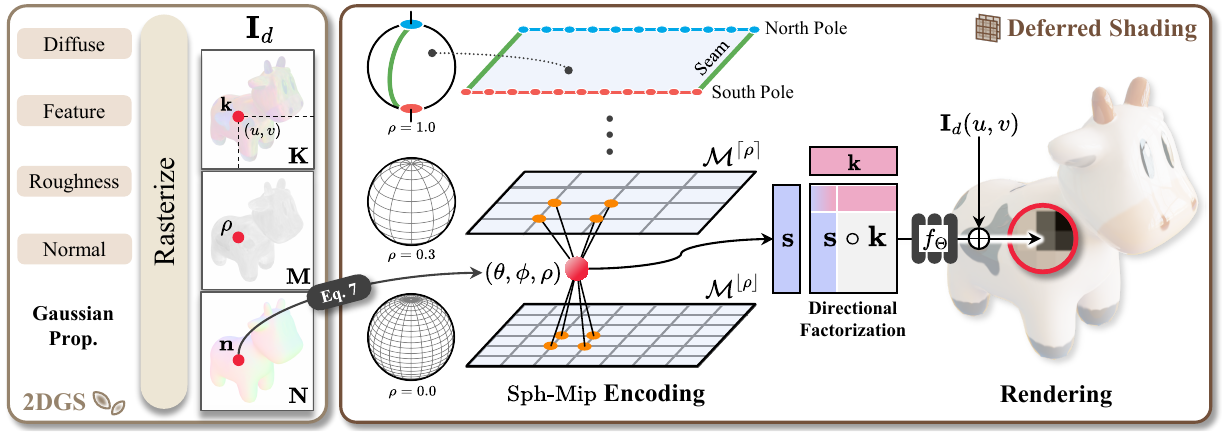}
\caption{Overview of \textit{Ref-GS}. From left to right: \textbf{the geometry pass} renders the scene properties, including appearance feature $\mathbf{K}$, roughness map $\mathbf{M}$, and normal map $\mathbf{N}$, into buffers via deferred rendering, \textbf{the lighting pass} projects the reflected direction $\omega_r$ onto spherical coordinates $(\theta, \phi)$ and featurized by $\operatorname{Sph-Mip}$ encoding for modeling far-field lighting, finally \textbf{the rendering pass} use tensor factorization $\mathbf{s} \circ \mathbf{k}$ to obtain spatially varying view-dependent effects and color each pixel $(u, v)$.
}
\label{fig:pipeline}

\end{figure*}

\section{Related Work}
\label{sec:Related}
Our work is closely related to research in novel view synthesis, and reflective and refractive object reconstruction and rendering.

\noindent \textbf{Novel View Synthesis.}
Novel view synthesis (NVS) focuses on generating novel views from a collection of posed images. A significant achievement in Neural Radiance Fields (NeRF)~\cite{mildenhall2020nerf} has been made for realistic NVS, thanks to implicit representations and volumetric rendering. Which has inspired other scene representations, including those for bounded objects~\cite{fridovich2022plenoxels,chen2022tensorf,sun2022direct,muller2022instant}, unbounded scenes~\cite{barron2022mipnerf360, kaizhang2020, martinbrualla2020nerfw, tancik2022block}, and scenes with high specular reflections and reflective effects~\cite{verbin2022refnerf,yao2022neilf,zhang2021nerfactor,Zhang_2023_ICCV}. Despite advancements, NeRF-based methods face challenges related to low training and rendering efficiency due to their implicit nature. Recently, 3D Gaussian Splatting (3DGS)\cite{kerbl3Dgaussians} has emerged as an alternative 3D representation to NeRF. While 3DGS achieves high-quality novel view synthesis, the reconstructed surface is generally noisy. To address the multi-view geometric inconsistencies in 3DGS, Huang \etal\cite{Huang2DGS2024} introduced 2D Gaussian Splatting (2DGS), where Gaussian disks are placed on object surfaces and smoothed locally. Our method extends 2DGS, significantly enhancing view-dependent effects and geometry quality.

\noindent \textbf{Reflective Scene Reconstruction and Rendering.}
Reflective scene reconstruction and rendering has been a challenging task, attracting significant attention. Directional encoding techniques have been explored to improve the modeling of reflections. Ref-NeRF~\cite{verbin2022refnerf} applies Integrated Directional Encoding (IDE) to enhance NeRF’s view-dependent effects but struggles with modeling near-field lighting. To address this limitation, Spec-NeRF~\cite{ma2024specnerf} introduces Gaussian Directional Encoding, improving the modeling of specular reflections under near-field lighting conditions. Wu \etal~\cite{wu2024neural} proposes Neural Directional Encoding, simulating near-field inter-reflections by tracking light cones within the NeRF model and utilizing a global cubemap filtered by a GGX kernel~\cite{walter2007microfacet} for reflection modeling. However, these methods rely on large multi-layer perceptrons (MLPs) to represent geometry, resulting in slower training and rendering speeds compared to Gaussian-based representations. Other approaches~\cite{liang2023envidr, liu2023nero, li2024tensosdf, yao2022neilf, zhang2023neilf++, jin2023tensoir} incorporate indirect lighting. ENVIDR~\cite{liang2023envidr} uses a neural renderer to learn physical light interactions through ray tracing, without explicitly formulating the rendering equation. NeRO~\cite{liu2023nero} introduces a lighting representation method using two MLPs and a split-sum approximation to model direct and indirect lighting, enabling high-quality reconstruction of reflective objects. However, NeRO requires extracting geometry from a pre-trained Signed Distance Function (SDF), which takes over 3 hours, resulting in significant inefficiencies. In contrast, recent Gaussian-based methods~\cite{jiang2024gaussianshader, wu2024deferredgs, tang20243igs, ye2024gsdr, zhu2024gs} offer more efficient solutions. For instance, 3iGS~\cite{tang20243igs} uses tensorial factorization~\cite{chen2022tensorf} to optimize incident illumination, while GaussianShader~\cite{jiang2024gaussianshader} separately models view-dependent effects, and 3DGS-DR~\cite{ye2024gsdr} incorporates deferred rendering for reflection modeling. While these methods excel in generating high-quality novel views, they still struggle to model near-field lighting, where environment maps change spatially. 

Moreover, concurrent work NU-NeRF~\cite{NU-NeRF} simulates reflection and refraction using physics-based ray tracing, improving the reconstruction of objects with fully transparent materials. To better demonstrate our method's capabilities, we evaluate reconstruction results on real-world scenes with transparent objects.
\section{Preliminaries}
\subsection{Gaussian Splatting}
Gaussian Splatting is a recent advance for efficient 3D reconstruction and rendering built upon rasterization. 3DGS and 2DGS are point-based representations that each point associated with geometry attributes (\ie $\mathbf{\Sigma} \in \mathbb{R}^{3\times3}$, position $\mu \in \mathbb{R}^3$ and opacity $\alpha$) and Spherical Harmonics (SH) appearance attributes $\mathbf{c}$, and the Gaussians are defined in world space centered at $\mu$:
\begin{equation}
    \label{eqn:gs-func}
    \mathcal{G}(\mathbf{x|\mu, \Sigma}) = \exp\left(-\frac{1}{2}(\mathbf{x - \mu})^T\mathbf{\Sigma}^{-1}(\mathbf{x - \mu})\right)
\end{equation}
where the covariance matrix is factorized into a rotation matrix $\bm{R}$ and a scaling matrix $\bm{S}$, to facilitate optimization:
\begin{equation}
\mathbf{\Sigma} = \bm{RSS^TR^T} 
\end{equation}
Note that, the surface of the 3D Gaussian is not well defined, leading to noisy surface reconstruction. To address this issue, 2D Gaussian Splatting (2DGS~\cite{Huang2DGS2024}) takes advantage of standard surfel modeling~\cite{pfister2000surfels, yifan2019differentiable, zwicker2001ewa} by adopting 2D oriented disks as surface elements and allows high-quality rendering with Gaussian splatting. Specifically, instead of evaluating a Gaussian’s value at the intersection between a pixel ray and a 3D Gaussian~\cite{kerbl3Dgaussians}, 2DGS evaluate Gaussian values at 2D disks and utilizes explicit ray-splat intersection, resulting in a perspective-correct splatting:
\begin{equation}
\mathcal{G}(\mathbf{u}) = \exp\left(-\frac{u(\mathbf{r})^2+v(\mathbf{r})^2}{2}\right)
\label{eq:gaussian-2d}
\end{equation}
where $\mathbf{u}=(u(\mathbf{r}),v(\mathbf{r}))$ is the intersection point between ray $\mathbf{r}$ and the primivitve in UV space. Furthermore, each Gaussian primitive has its own view-dependent color $\mathbf{c}$ with SH coefficients. For rendering, Gaussians are sorted according to their centers and composed into pixels with front-to-back alpha blending:
\begin{equation}
\label{2dgs}
\mathbf{c}(\mathbf{r}) = \sum_{i=1} \mathbf{c}_i \alpha_i \mathcal{G}_i(\mathbf{u}) T_i
\end{equation}
where $T_i$ is approximated accumulated transmittances defined by $\prod_{j=1}^{i-1} (1 - \alpha_j \mathcal{G}_j(\mathbf{u}))$.
Note that both 3DGS and 2DGS are forward processes, where scenes are directly projected onto the image plane. Each Gaussian primitive is rendered and lighted in object space before being mapped to screen space. However, forward rendering generally tends to waste a lot of fragment shader runs in scenes with a high depth complexity (multiple primitives cover the same screen pixel) as fragment shader outputs are overwritten.

\subsection{Deferred Shading}
In 3D computer graphics, deferred shading~\cite{Deering_Winner_Schediwy_Duffy_Hunt} is a screen-space shading technique designed to significantly reduce the number of shading operations compared to the forward rendering process.

Deferred shading is a technique that defers most intensive rendering operations, such as lighting calculations, to a later stage in the rendering pipeline. This technique involves two main passes. In the first pass, known as the geometry pass, the scene is rendered once to capture various types of geometric information from objects in the scene. These data are stored in a collection of textures called the G-buffer, which contains information such as position vectors, color vectors, normal vectors, and specular values. The G-buffer thus serves as a repository of scene geometry that can be utilized for subsequent, potentially complex, lighting calculations.

In the second pass, referred to as the lighting pass, the G-buffer textures are used to calculate lighting across the scene. A screen-filled quad is rendered, and the lighting for each fragment is computed using the geometric information stored in the G-buffer, iterating over each pixel. This process decouples advanced fragment processing from the initial rendering of each object, allowing lighting calculations to draw directly from the G-buffer textures rather than the vertex shader, with additional input from uniform variables as needed. This allows to maintain the same lighting calculations but optimizes the process by postponing them until after the G-buffer has been populated.

\begin{figure}[t]
\centering
\includegraphics[width=1.0\linewidth]{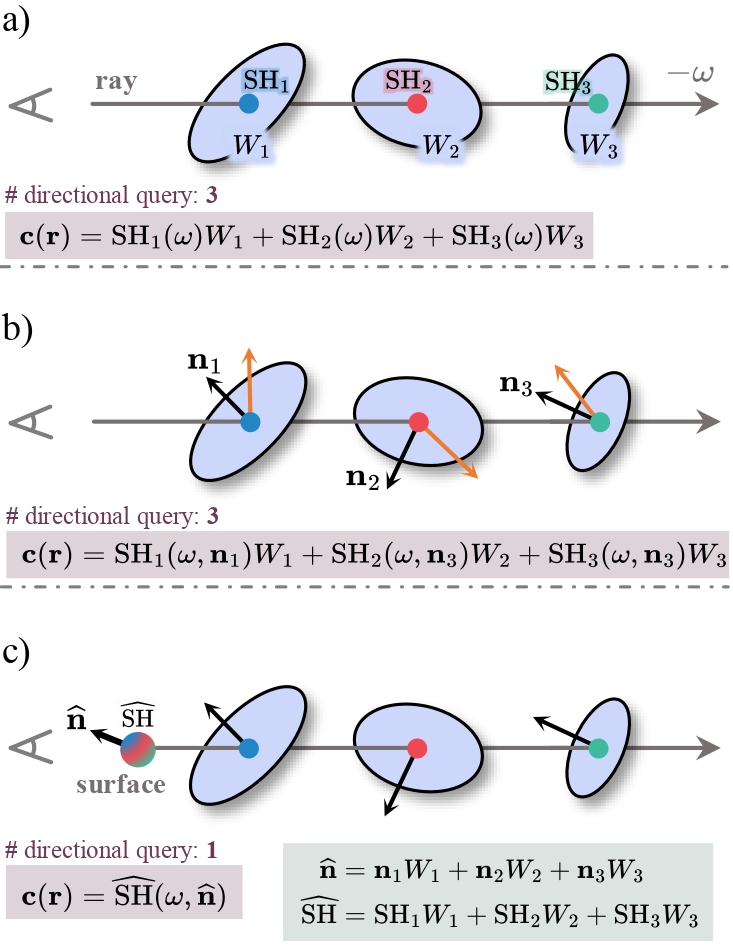}
\caption{\textbf{Comparison of directional query in Gaussian Splatting.} a) The original 3DGS~\cite{kerbl3Dgaussians} and 2DGS~\cite{Huang2DGS2024} methods query each primitive's SH coefficient using the viewing direction, then accumulate view-dependent radiance as the ray color. b) Ref-NeRF~\cite{verbin2022refnerf} and recent GaussianShader~\cite{jiang2024gaussianshader} utilize the reflection direction transformed by both the viewing and normal directions as the directional query. c) We introduce Gaussian deferred shading by first integrating the SH coefficient and normal as a surface point, then evaluating its view-dependent color.}
\label{fig:ambiguty}
\end{figure}

\section{Ambiguity in Directional Query} 
\label{sec:ambiguty}
In prior Gaussian Splatting methods, the diffusion, reflection and refraction components of each primitive are simplified using view-dependent emission radiance, which significantly accelerates the forward rendering process without the need for per-instance lighting evaluation. Then, they optimize the emission radiance together with the geometry jointly through an inverse rendering framework by backpropagating a multi-view photometric loss.

We observe that this modeling suffers from serious representation ambiguities as illustrated in Fig.~\ref{fig:ambiguty} shows an integration process of three primitives. Consider the integration processing used by vanilla 3D and 2D Gaussian Splatting in (a), it queries view-dependent color using viewing direction, leading to strong bias to diffuse materials and high-frequency irradiance is generally fake by complex primitive overlaying; to handle strong reflection, Ref-NeRF~\cite{verbin2022refnerf} and its follow-ups~\cite{liang2023envidr, liu2023nero} utilize reflection direction as color query by considering point normal in (b). Unlike the continuous representation in NeRF that neighboring points' attributes are regularized to each other, Gaussian Splatting treats each point independently. Directly applying reflection direction provides limited gains due to the ambiguity between SH coefficients and the primitive orientation, \ie transforming viewing direction to reflection direction can be eliminated by the changing in SH coefficient. In practice, since irradiance is independent across primitives and multiple primitives contribute to a target ray, this inherently introduces strong ambiguities, leading to noisy reconstructions.

\section{Ref-GS}
Our approach aims to reconstruct photorealistic view-dependent effect. The overview of our method is shown in Fig.~\ref{fig:pipeline}. Specifically, we present a deferred Gaussian splatting to generate a G-buffer (Section~\ref{Sec: Deferred Gaussian}). We then introduce a directional factorization for representing spatially varying view-dependent effects (Section~\ref{Sec: Directional Factorization}) and a multi-level spherical feature grid that models far-field lighting (Section~\ref{Sec: Far-field Lighting}).

\label{sec:method}
\subsection{Deferred Gaussian}
\label{Sec: Deferred Gaussian}

We now introduce a novel deferred Gaussian Splatting method to address the ambiguity issue discussed in Section~\ref{sec:ambiguty}. 
Direct volume integration of Gaussian representations can result in blurry view-dependent effects and noisy surfaces due to ambiguity in directional queries. Our solution is to first blend Gaussian attributes, then apply shading, similar to deferred shading.
To be specific, we perform alpha blending on primitive attributes (\ie, for the $i^{th}$ Gaussian include diffuse color ${\mathbf{c}_d}_i \in \mathbb{R}^3$, feature $\mathbf{f}_{i} \in \mathbb{R}^D$, roughness $\rho_i \in [0, 1]$) along the rays and convert the attributes into color in image space, as described in Eq.~\ref{2dgs} and (c) of Fig.~\ref{fig:ambiguty}. Additionally, the color of each pixel is decomposed into a diffuse component $\mathbf{I}_d$ and a specular component, queried by the reflected direction $\omega_r \in \mathbb{R}^3$ with surface normal $\mathbf{n} \in \mathbb{R}^3$.
We use the integrated diffuse color $\mathbf{I}_d$ directly as the ray's diffuse component and obtain view-dependent effects at each pixel through a shader $f_\Theta$, conditioned on the spatial feature $\mathbf{K} \in \mathbb{R}^{H \times W \times D}$ and the directional feature $\mathbf{S} \in \mathbb{R}^{H \times W \times C}$:
\begin{equation}
\label{deferredShading}
\mathbf{I} = \mathbf{I}_d + f_\Theta(\mathbf{S}, \mathbf{K} \otimes \mathbf{S})
\end{equation}
where $\otimes$ denotes the per-pixel outer product, obtaining the high-dimensional intermediate tensor with the shape of $H \times W \times (D \times C)$. 

Note that the feature $\mathbf{K}$ represents the expected feature of each pixel and is obtained by splatting per-primitive features $\mathbf{f}_{i}$ using Eq.~\ref{2dgs}. Similarly, we generate the roughness map $\mathbf{M}$ corresponding to $\rho_i$ and the normal map $\mathbf{N}$. In practice, we treat ${\mathbf{M}, \mathbf{N}, \mathbf{K}}$ as a G-buffer and pass it a standard rasterization render for shading.

\subsection{Directional Factorization}
\label{Sec: Directional Factorization}
In essence, the key to modeling view-dependent effects is accurately capturing spatially varying near- and far-field inter-reflections.
Prior methods~\cite{jiang2024gaussianshader, ye2024gsdr, nvdiffrec_2022_CVPR} often rely on a global 2D environment map for far-field lighting, assuming all light sources are at an infinite distance. Other methods~\cite{verbin2022refnerf, tang20243igs} only model direct lighting. These assumptions are insufficient for reconstructing surfaces under near-field lighting, especially in scenes where light sources or objects are close to the target object.

Inspired by TensoRF~\cite{chen2022tensorf}, we propose a low-rank tensor factorization $\mathbf{s} \circ \mathbf{k}$ to represent spatio-angular view-dependent effects, where $\circ$ denotes the outer product. 
As illustrated in Fig.~\ref{fig:pipeline}, we connect the spatial feature vector $\mathbf{k} \in \mathbb{R}^D$ and directional feature vector $\mathbf{s} \in \mathbb{R}^C$ using a simple vector outer product to form a block matrix, which is then flattened into a 1D vector and fed into a lightweight MLP decoder for final color prediction.
The outer product of spatial feature $\mathbf{k}$ and directional feature $\mathbf{s}$ enables decomposition of geometry and lighting, while effectively capturing essential information such as global lighting, shadows, and self-occlusion.

Our factorization-based model is simple yet effective for representing spatially varying view-dependent effects in complex reflective scenes, enhancing both novel view synthesis and surface reconstruction. Additionally, this factorization reduces the feature channels for each Gaussian primitive, significantly lowering the computational overhead in volume rendering and scene representation.

\begin{figure}[t]
\centering
\includegraphics[width=1.0\linewidth]{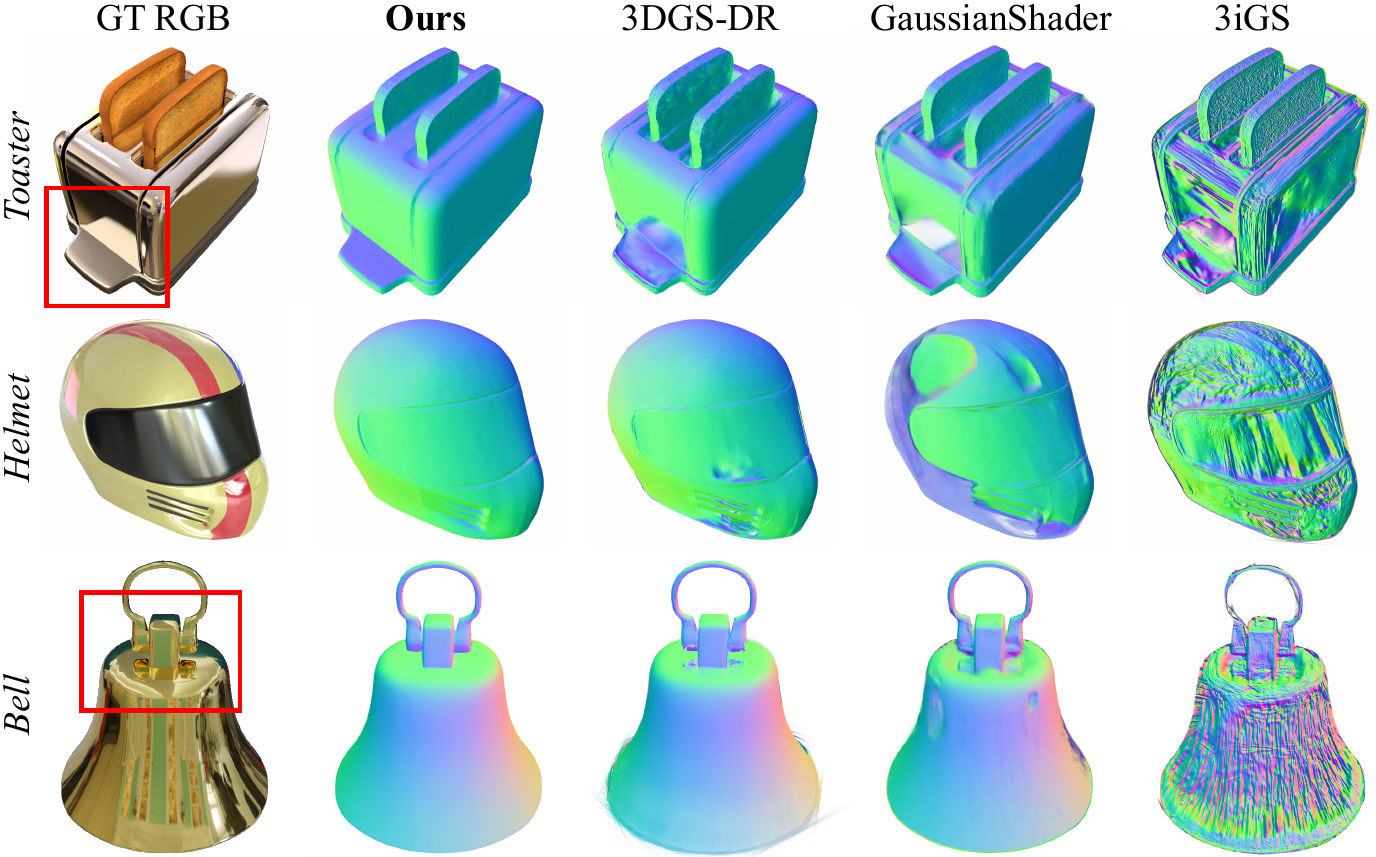}
\caption{\textbf{Visualized estimated surface normal results synthetic datasets}~\cite{verbin2022refnerf, liu2023nero}. Compared to existing Gaussian-based methods, our method has more accurate surface reconstruction for shiny objects with  inter-reflections, as depicted for this \textit{`Toaster'} and \textit{`Bell'} scenes.
 }
\label{fig:synthetic-Normal}
\end{figure}

\begin{figure*}[t]
\vspace{-6pt}
\centering
\includegraphics[width=1.0\linewidth]{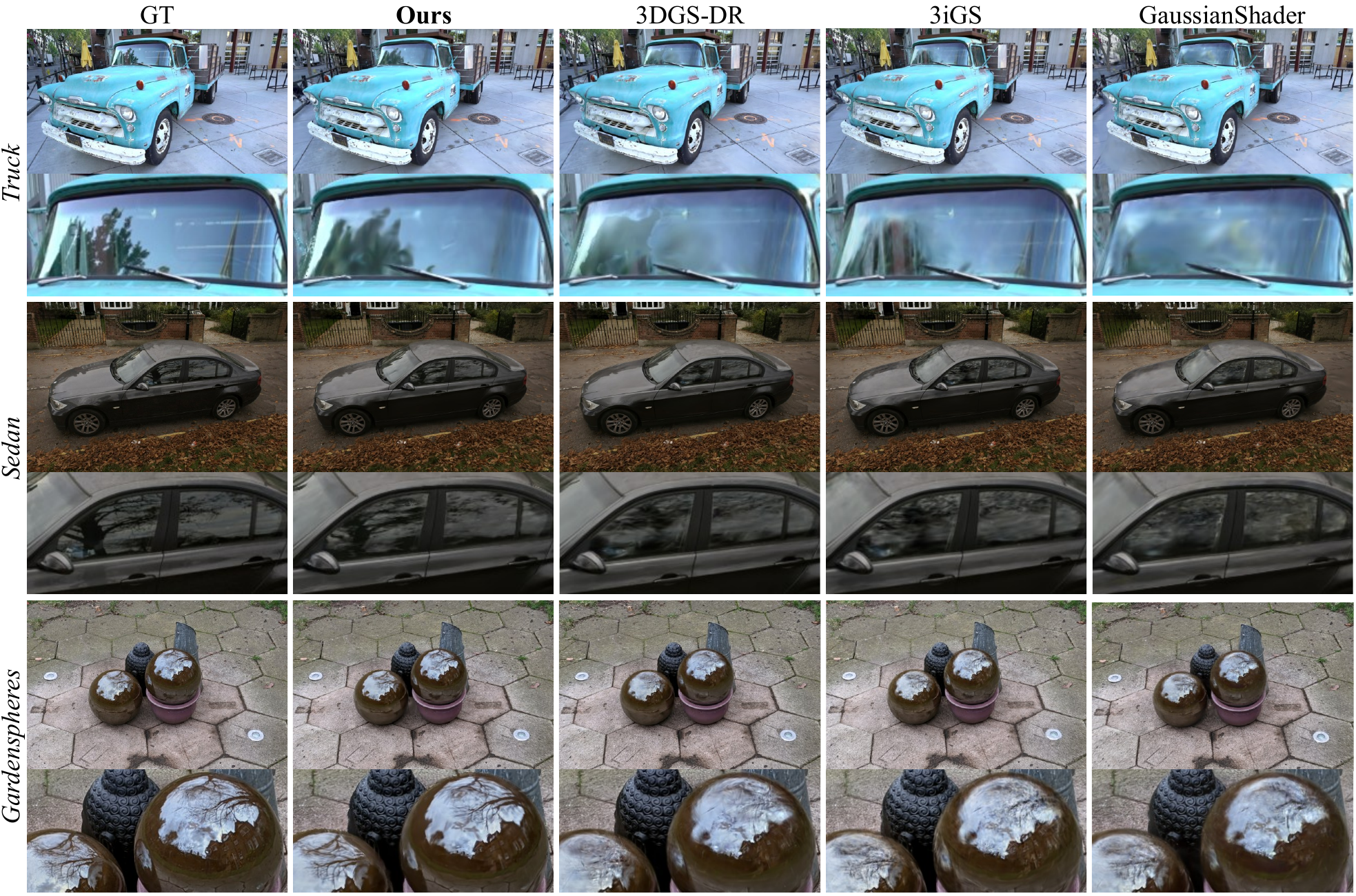}
\caption{\textbf{Qualitative comparisons of test-set views of real-world scenes.} Notice the high-frequency reflections rendered by our model, including sharp details of the tree branches and buildings reflected in the sphere.}
\label{fig:real}
\end{figure*}

\subsection{Far-field Lighting}
\label{Sec: Far-field Lighting}
We now present a novel $\operatorname{Sph-Mip}$ encoding for modeling high-frequency far-field lighting, using a learnable multi-level spherical feature grid, named $\operatorname{Sph-Mip}$ grid.

We utilize a longitude-latitude lattice (\textit{Long.-Lat.}) to distribute feature points on a spherical surface and unfold them into a 2D feature grid for efficient indexing.
Given the G-buffer $\{\mathbf{M}, \mathbf{N}, \mathbf{K}\}$, the normal $\mathbf{n}=\mathbf{N}(u, v)$, roughness $\rho=\mathbf{M}(u, v)$ for the pixel $(u, v)$, we have:

\begin{equation}
\mathbf{s} = \operatorname{Sph-Mip}(\omega_r, \rho, \mathcal{M})
\end{equation}
where  $\omega_r$ denotes the reflection direction reflected by the surface normal $\mathbf{n}$ and viewing direction $\omega_i$.

Note that, as shown in Fig.~\ref{fig:pipeline}, the $\operatorname{Sph-Mip}$ grid is three-dimensional, the directional coordinates $(\theta, \phi)$ correspond to the XY axes of the grid, while the Z-axis represents the roughness $\rho$ variance. Given rasterized buffers, we first calculate its corresponding spherical coordinates $(\theta, \phi)$ for each pixel by:
\begin{equation}
    \begin{aligned}
    \theta &= \text{arccos} \left(\frac{\omega_r^z}{\sqrt{{\omega_r^x}^2+{\omega_r^y}^2+{\omega_r^z}^2}}\right)  \in \left [0, \pi \right ] \\
    \phi &= \text{arctan2} \left(\frac{{\omega_r^y}}{{\omega_r^x}}\right)  \in \left [-\pi, \pi \right ]
    \end{aligned}
\end{equation}

Then, given the surface roughness $\rho$, we interpolate features along the roughness dimension. In practice, we resize the grids at different levels to the same resolution during the feature query, facilitating efficient three-dimensional interpolation using trilinear interpolation with the coordinates $(\theta, \phi, \rho)$ in the fragment shader.

For the mipmap resolution, we define the base level $\mathcal{M}^{L_0}$ at the highest resolution of $H_\mathcal{M} \times W_\mathcal{M} \times C$, where $H_\mathcal{M}$, $W_\mathcal{M}$, and $C$ represent the height, width, and number of channels, respectively.
While the resolution for other levels ($\mathcal{M}^{L_i}, i = {1, 2, \dots, N}$) is divided by $2 \times$ along the height and width dimensions.

\section{Experiments and Results}

\begin{table}[t]
    \centering
    \resizebox{0.48\textwidth}{!}{ 

    \begin{tabular}{lccccccc|cccc}
    \hline

    \multicolumn{1}{c}{ } & \multicolumn{7}{c|}{Shiny Blender} & \multicolumn{4}{c}{Shiny Real}\\
    \multicolumn{1}{c}{ } & Car & Ball & Helmet & Teapot & Toaster & Coffee & Avg. & Garden & Sedan & Toycar & Avg.\\ \hline     
             
    \multicolumn{12}{c}{PSNR$\uparrow$} \\ \hline
    Ref-NeRF~\cite{verbin2022refnerf} & \cellcolor{yellow!25}30.41 & 29.14 & 29.92 & 45.19 & 25.29 & \cellcolor{yellow!25}33.99 & 32.32 & \cellcolor{orange!25}22.01 & 25.21 & 23.65 & 23.62 
 \\
    NeRO~\cite{liu2023nero} & 25.53 & 30.26 & 29.20 & 38.70 & \cellcolor{yellow!25}26.46 & 28.89 & 29.84 & — & — & — & — 
 \\
    ENVIDR~\cite{liang2023envidr} & 28.46 & \cellcolor{red!25}38.89 & \cellcolor{orange!25}32.73 & 41.59 & 26.11 & 29.48 & \cellcolor{yellow!25}32.88 & 21.47 & 24.61 & 22.92 & 23.00 
 \\
    3DGS~\cite{kerbl3Dgaussians} & 27.24 & 27.69 & 28.32 & 45.68 & 20.99 & 32.32 & 30.37 & 21.75 & 26.03 & \cellcolor{orange!25}23.78 & \cellcolor{yellow!25}23.85 
 \\
    GaussianShader~\cite{jiang2024gaussianshader} & 27.51 & 29.02 & 28.73 & 43.05 & 22.86 & 31.34 & 30.42 & 21.74 & 24.89 & \cellcolor{yellow!25}23.76 & 23.46 
 \\
    3iGS~\cite{tang20243igs} & 27.52 & 26.82 & 28.08 & \cellcolor{yellow!25}46.05 & 22.71 & 32.64 & 30.64 & \cellcolor{yellow!25}21.96 & \cellcolor{orange!25}26.59 & 23.75 & \cellcolor{orange!25}24.10 
 \\
    3DGS-DR~\cite{ye2024gsdr} & \cellcolor{orange!25}30.43 & \cellcolor{yellow!25}33.44 & \cellcolor{yellow!25}31.49 & \cellcolor{red!25}47.00 & \cellcolor{orange!25}26.69 & \cellcolor{red!25}34.61 & \cellcolor{orange!25}33.94 & 21.52 & \cellcolor{yellow!25}26.32 & 23.57 & 23.80 
 \\
    Ours & \cellcolor{red!25}30.94 & \cellcolor{orange!25}36.10 & \cellcolor{red!25}33.40 & \cellcolor{orange!25}46.69 & \cellcolor{red!25}27.28 & \cellcolor{orange!25}34.38 & \cellcolor{red!25}34.80 & \cellcolor{red!25}22.48 & \cellcolor{red!25}26.63 & \cellcolor{red!25}24.20 & \cellcolor{red!25}24.44 
 \\ \hline
 
    \multicolumn{12}{c}{SSIM$\uparrow$} \\ \hline
    Ref-NeRF~\cite{verbin2022refnerf} & \cellcolor{yellow!25}0.949 & 0.956 & 0.955 & \cellcolor{yellow!25}0.995 & 0.910 & \cellcolor{yellow!25}0.972 & 0.956 & \cellcolor{orange!25}0.584 & 0.720 & 0.633 & 0.646 
 \\
    NeRO~\cite{liu2023nero} & \cellcolor{yellow!25}0.949 & 0.974 & \cellcolor{yellow!25}0.971 & \cellcolor{yellow!25}0.995 & 0.929 & 0.956 & 0.962 & — & — & — & —
 \\
    ENVIDR~\cite{liang2023envidr} & \cellcolor{orange!25}0.961 & \cellcolor{red!25}0.991 & \cellcolor{red!25}0.980 & \cellcolor{orange!25}0.996 & \cellcolor{yellow!25}0.939 & 0.949 & \cellcolor{yellow!25}0.969 & 0.561 & 0.707 & 0.549 & 0.606 
 \\   
    3DGS~\cite{kerbl3Dgaussians} & 0.930 & 0.937 & 0.951 & \cellcolor{orange!25}0.996 & 0.895 & 0.971 & 0.947 & 0.571 & 0.771 & \cellcolor{orange!25}0.637 & \cellcolor{orange!25}0.660 
 \\
    GaussianShader~\cite{jiang2024gaussianshader} & 0.930 & 0.954 & 0.955 & \cellcolor{yellow!25}0.995 & 0.900 & 0.969 & 0.951 & \cellcolor{yellow!25}0.576 & 0.728 & \cellcolor{orange!25}0.637 & 0.647 
 \\
    3iGS~\cite{tang20243igs} & 0.930 & 0.933 & 0.951 & \cellcolor{orange!25}0.996 & 0.909 & \cellcolor{yellow!25}0.972 & 0.948 & 0.557 & \cellcolor{red!25}0.789 & 0.626 & 0.657 
 \\
    3DGS-DR~\cite{ye2024gsdr} & \cellcolor{red!25}0.962 & \cellcolor{yellow!25}0.979 & \cellcolor{yellow!25}0.971 & \cellcolor{red!25}0.997 & \cellcolor{orange!25}0.942 & \cellcolor{red!25}0.976 & \cellcolor{orange!25}0.971 & 0.570 & \cellcolor{yellow!25}0.773 & \cellcolor{yellow!25}0.635 & \cellcolor{yellow!25}0.659 
 \\
    Ours & \cellcolor{orange!25}0.961 & \cellcolor{orange!25}0.981 & \cellcolor{orange!25}0.975 & \cellcolor{red!25}0.997 & \cellcolor{red!25}0.950 & \cellcolor{orange!25}0.973 & \cellcolor{red!25}0.973 & \cellcolor{red!25}0.607 & \cellcolor{orange!25}0.783 & \cellcolor{red!25}0.656 & \cellcolor{red!25}0.682 
 \\ \hline

    \multicolumn{12}{c}{LPIPS$\downarrow$} \\ \hline
    Ref-NeRF~\cite{verbin2022refnerf} & 0.051 & 0.307 & 0.087 & 0.013 & 0.118 & 0.082 & 0.110 & \cellcolor{yellow!25}0.251 & 0.234 & \cellcolor{red!25}0.231 & 0.239 
 \\
    NeRO~\cite{liu2023nero} & 0.074 & \cellcolor{orange!25}0.094 & \cellcolor{orange!25}0.050 & 0.012 & \cellcolor{yellow!25}0.089 & 0.110 & \cellcolor{yellow!25}0.072 & — & — & — & — 
 \\
    ENVIDR~\cite{liang2023envidr} & 0.049 & \cellcolor{red!25}0.067 & \cellcolor{yellow!25}0.051 & \cellcolor{yellow!25}0.011 & 0.116 & 0.139 & \cellcolor{yellow!25}0.072 & 0.263 & 0.387 & 0.345 & 0.332 
 \\
    3DGS~\cite{kerbl3Dgaussians} & \cellcolor{yellow!25}0.047 & 0.161 & 0.079 & \cellcolor{orange!25}0.007 & 0.126 & \cellcolor{yellow!25}0.078 & 0.083 & \cellcolor{orange!25}0.248 & \cellcolor{yellow!25}0.206 & \cellcolor{yellow!25}0.237 & \cellcolor{orange!25}0.230 
 \\
    GaussianShader~\cite{jiang2024gaussianshader} & \cellcolor{orange!25}0.045 & 0.148 & 0.088 & 0.012 & 0.111 & 0.085 & 0.082 & 0.274 & 0.259 & 0.239 & 0.257
 \\
    3iGS~\cite{tang20243igs} & \cellcolor{orange!25}0.045 & 0.166 & 0.073 & \cellcolor{red!25}0.006 & 0.098 & \cellcolor{orange!25}0.077 & 0.077 & 0.252 & \cellcolor{red!25}0.190 & 0.251 & \cellcolor{yellow!25}0.231 
 \\
    3DGS-DR~\cite{ye2024gsdr} & \cellcolor{red!25}0.034 & 0.104 & \cellcolor{orange!25}0.050 & \cellcolor{red!25}0.006 & \cellcolor{orange!25}0.083 & \cellcolor{red!25}0.076 & \cellcolor{orange!25}0.059 & \cellcolor{yellow!25}0.251 & 0.208 & 0.249 & 0.236 
 \\ 
    Ours & \cellcolor{red!25}0.034 & \cellcolor{yellow!25}0.098 & \cellcolor{red!25}0.045 & \cellcolor{red!25}0.006 & \cellcolor{red!25}0.070 & 0.082 & \cellcolor{red!25}0.056 & \cellcolor{red!25}0.242 & \cellcolor{orange!25}0.196 & \cellcolor{orange!25}0.236 & \cellcolor{red!25}0.224 
 \\ \hline
\end{tabular}
}

\caption{
\textbf{Quantitative NVS comparisons on Shiny Blender and Shiny Real datasets}~\cite{verbin2022refnerf}. Our method is comparable with both Gaussian-based methods and prior reflective object reconstruction methods. \textit{`Gardenspheres'} is abbreviated as \textit{`Garden'}.
}
\label{table:shiny blender real}
\end{table}

\subsection{Datasets}
We evaluate our method on several synthetic and real-world datasets. For the synthetic datasets, we evaluate our model on NeRF Synthetic~\cite{mildenhall2020nerf}, which contains scenes of complex
geometries with realistic non-Lambertian materials. Similarly, we evaluate our model on reflective objects using Shiny Blender~\cite{verbin2022refnerf} and Glossy Synthetic~\cite{liu2023nero}. For the real-world datasets, we use Shiny Real dataset captured from~\cite{verbin2022refnerf}, as well as scenes with reflections from Mip-NeRF360~\cite{barron2022mipnerf360} and Tanks $\&$ Temples~\cite{knapitsch2017tanks}. Additionally, we use Glass $\&$ Ball~\cite{bemana2022eikonal}, which contains refractive objects with unknown geometry, to show the generalization ability of our method for diverse materials.

\begin{figure}[t]
\centering
\includegraphics[width=1.0\linewidth]{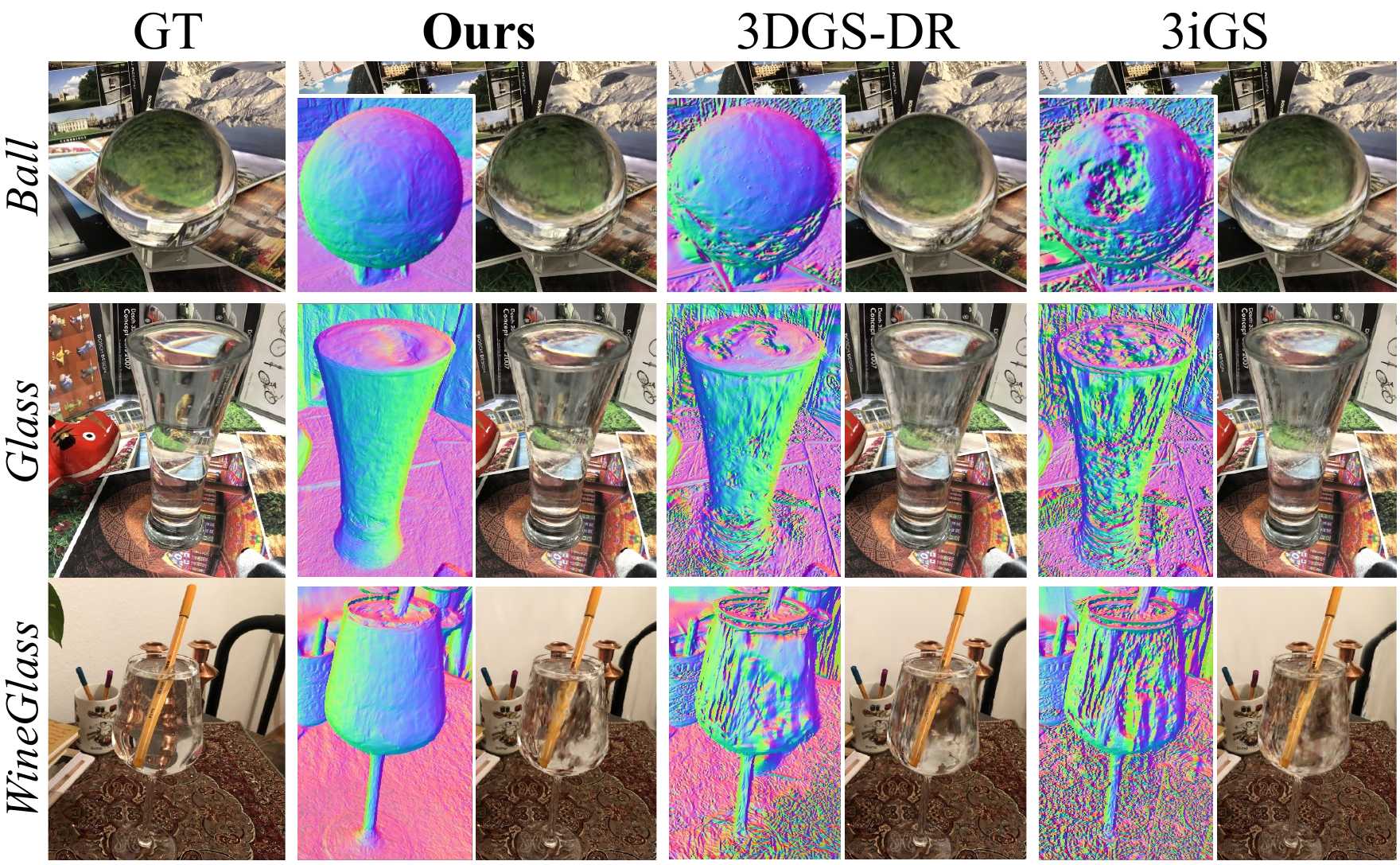}
\caption{ \textbf{Comparison results on refractive scenes.}
Normal reconstruction and rendering results on real scenes from Glass $\&$ Ball~\cite{bemana2022eikonal}. Our method performs significantly better than 3DGS-DR~\cite{ye2024gsdr} and 3iGS~\cite{tang20243igs}.
}
\label{fig:real-glass}
\end{figure}

\begin{figure*}[t]
\centering
\includegraphics[width=1.0\linewidth]{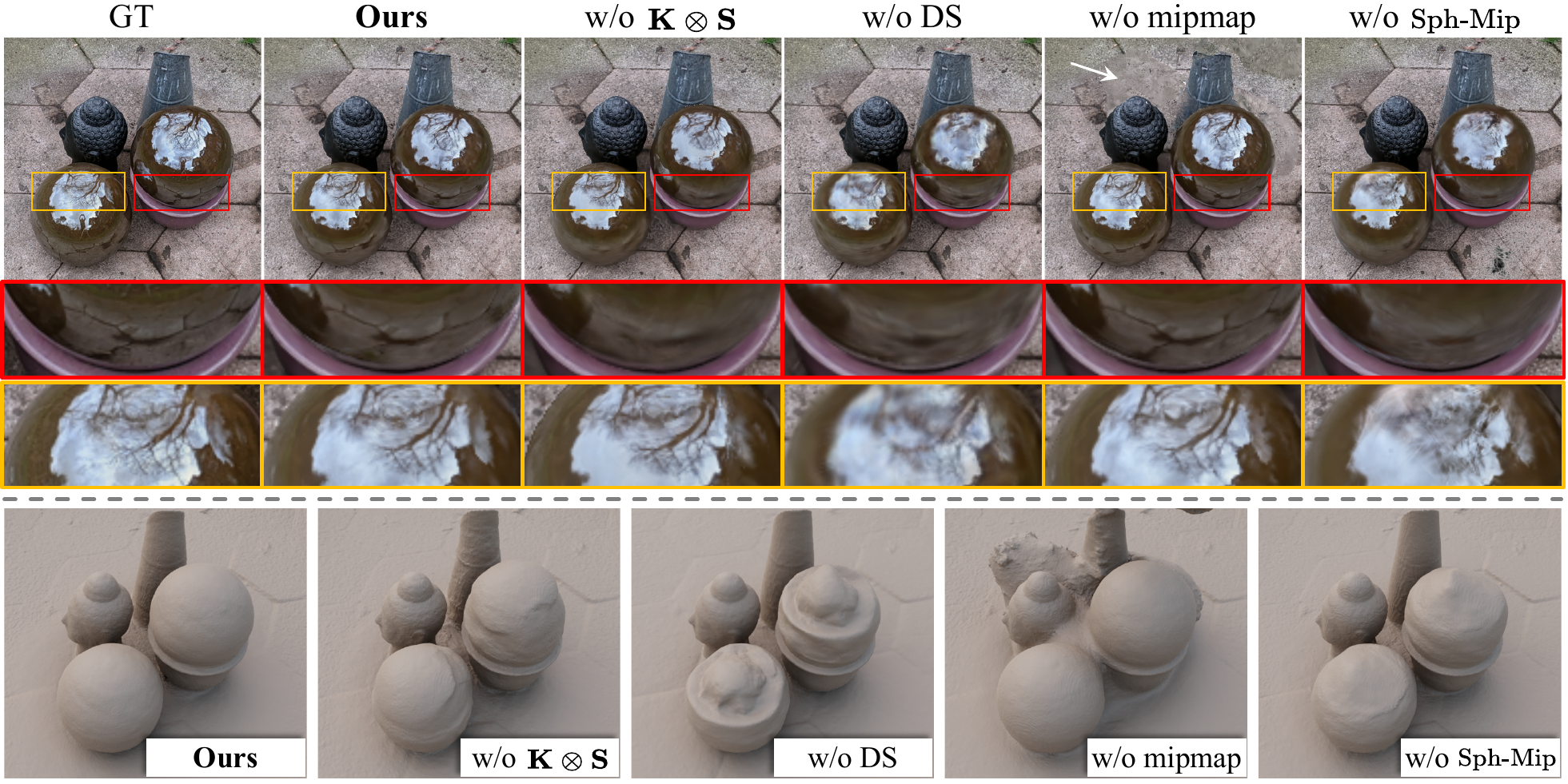}
\caption{\textbf{Qualitative ablation on the \textit{`Gardenspheres'} scene~\cite{verbin2022refnerf}.} Using the G-buffer instead of $\operatorname{Sph-Mip}$ (\ie, w/o $\operatorname{Sph-Mip}$) or without deferred shading (\ie, w/o DS), sharp details, such as tree branches reflected in the sphere, are not accurately reconstructed. It is necessary to use multi-level spherical feature grid strategies (\ie, w/o mipmap), otherwise rough surfaces will fail to be reconstructed and artifacts will appear during rendering. Additionally, directional factorization (\ie, w/o $\mathbf{K} \otimes \mathbf{S}$) is essential for modeling near-field inter-reflections.}
\label{fig:Ablation study}
\end{figure*}

\subsection{Baselines and metrics}
We compare our method with the following baselines: Ref-NeRF~\cite{verbin2022refnerf}, a NeRF-based method focusing on reflective objects rendering; SDF-based methods including ENVIDR~\cite{liang2023envidr} and NeRO~\cite{liu2023nero}, top-performing implicit methods for reconstructing reflective objects; and Gaussian-based methods such as GaussianShader~\cite{jiang2024gaussianshader}, 3iGS~\cite{tang20243igs} and 3DGS-DR~\cite{ye2024gsdr}. We trained these models based on their public codes and configurations. Evaluation metrics for rendering quality include PSNR, SSIM~\cite{Wang_Bovik_Sheikh_Simoncelli_2004}, and LPIPS~\cite{Zhang_Isola_Efros_Shechtman_Wang_2018}. Additionally, we use Mean Angular Error in degrees (MAE$^\circ$) to evaluate the normal accuracy.

\subsection{Implementation Details}
All experiments are conducted on a single Tesla V100 GPU with 32GB of VRAM. The parameters to be optimized include the MLP $f_\Theta$, the mipmap $\mathcal{M}$ and each 2D Gaussian's parameters (\textit{e.g.} the feature $\mathbf{f}_{i} \in \mathbb{R}^4$).
Following the approach in~\cite{Huang2DGS2024}, we optimize these parameters using differentiable splatting and gradient-based backpropagation. Optimization was performed over 30,000 iterations using the Adam optimizer~\cite{kingma2014adam}.
We implement our $\operatorname{Sph-Mip}$ using PyTorch~\cite{2019PyTorch} framework, and employ the Nvdiffrast~\cite{Laine_Hellsten_Karras_Seol_Lehtinen_Aila_2020} library for efficient mipmap querying. The shape of the base level of the mipmap $\mathcal{M}^{L_0}$ in $\operatorname{Sph-Mip}$ encoding is empirically set to $H_\mathcal{M}=512, W_\mathcal{M}=1024, C=16$, and the number of levels is $N=9$. For our implicit representation of the specluar color prediction, we use a lightweight MLP with 1 hidden layers of size 256. We use the ReLU activation function. We propose to train our model with the same loss function $\mathcal{L}$ as 2DGS~\cite{Huang2DGS2024}.

\subsection{Comparisons}
Quantitative results on synthetic datasets are reported in Tab.~\ref{table:shiny blender real} and Tab.~\ref{table:NeRF+Glossy}, where high-quality reflection modeling relies on accurate normal estimation, as shown in Fig.~\ref{fig:synthetic-Normal}.
Furthermore, we report the training and rendering speeds (tested) of our model on the same hardware in Tab.~\ref{table:speed}, comparing it with existing Gaussian-based methods. Our method achieves a balance between quality and training speed. Although the speed of our model is not as fast as 3DGS~\cite{kerbl3Dgaussians}, it remains competitive and achieves real-time rendering speeds.

To demonstrate the effectiveness of our method in real-world scenes, rather than just small objects, we evaluated our renderings on the Shiny Real dataset from Ref-NeRF~\cite{verbin2022refnerf} as shown in Tab.~\ref{table:shiny blender real} and Fig.~\ref{fig:real}. Furthermore, the qualitative results in Fig.~\ref{fig:real-glass} show that for real-world scenes with refractive objects, our model outperforms 3iGS~\cite{tang20243igs} and 3DGS-DR~\cite{ye2024gsdr} on the Glass $\&$ Ball dataset from Eikonal Fields~\cite{bemana2022eikonal}.

\subsection{Ablation Studies} \label{sec:ablation}

\begin{table}[t]
    \Huge
    \centering
    \resizebox{0.48\textwidth}{!}{ 
    \begin{tabular}{lccc|ccc|c}
    \hline
    \multicolumn{1}{c}{ } & \multicolumn{3}{c|}{NeRF Synthetic} & \multicolumn{3}{c|}{Glossy Synthetic} & \multicolumn{1}{c}{ShinyB}\\
             & PSNR$\uparrow$ & SSIM$\uparrow$ & LPIPS$\downarrow$ & PSNR$\uparrow$ & SSIM$\uparrow$ & LPIPS$\downarrow$ & MAE$^\circ$$\downarrow$
 \\ \hline 
    Ref-NeRF~\cite{verbin2022refnerf} & 31.29 & 0.947 & 0.058 & 27.50 & 0.927 & 0.100 & 18.38 
 \\
    ENVIDR~\cite{liang2023envidr} & 28.13 & 0.953 & 0.068 & 29.58 & 0.952 & \cellcolor{red!25}0.057 & \cellcolor{yellow!25}4.61 
 \\
    3DGS~\cite{kerbl3Dgaussians} & \cellcolor{orange!25}33.30 & \cellcolor{orange!25}0.969 & \cellcolor{orange!25}0.030 & 26.50 & 0.917 & 0.092 & —
 \\
    GShader~\cite{jiang2024gaussianshader} & 31.48 & 0.960 & 0.042 & 27.54 & 0.922 & \cellcolor{yellow!25}0.087 & 10.93 
 \\
    3iGS~\cite{tang20243igs} & \cellcolor{red!25}33.60 & \cellcolor{red!25}0.970 & \cellcolor{red!25}0.029 & 26.39 & 0.913 & 0.089 & 15.47 
 \\
    3DGS-DR~\cite{ye2024gsdr} & 31.02 & 0.962 & 0.047 & \cellcolor{orange!25}29.78 & \cellcolor{yellow!25}0.954 & \cellcolor{red!25}0.057 & \cellcolor{orange!25}2.43 
 \\
    GS-ROR~\cite{zhu2024gs} & — & — & — & \cellcolor{yellow!25}29.70 & \cellcolor{orange!25}0.956 & — & 7.23 
 \\
    Ours & \cellcolor{yellow!25}33.20 & \cellcolor{yellow!25}0.966 & \cellcolor{yellow!25}0.036 & \cellcolor{red!25}30.59 & \cellcolor{red!25}0.957 & \cellcolor{orange!25}0.058 & \cellcolor{red!25}2.21 
 \\ \hline
\end{tabular}
}
\caption{\textbf{Quantitative NVS comparisons on NeRF Synthetic~\cite{mildenhall2020nerf} and Glossy Synthetic~\cite{liu2023nero} datasets.} Normal reconstruction quality on the Shiny Blender (ShinyB)~\cite{verbin2022refnerf} dataset evaluated by MAE$^\circ$. GaussianShader~\cite{jiang2024gaussianshader} is abbreviated as GShader.}
\label{table:NeRF+Glossy}
\end{table}

\begin{table}[t]
    \centering
    \resizebox{0.43\textwidth}{!}{ 

    \begin{tabular}{lc|c|c|c}
    \multicolumn{1}{c}{ } & \multicolumn{1}{c|}{PSNR$\uparrow$} & \multicolumn{1}{c|}{SSIM$\uparrow$} & \multicolumn{1}{c|}{LPIPS$\downarrow$} & \multicolumn{1}{c}{MAE$^\circ$$\downarrow$}
 \\ \hline
    w/o $\operatorname{Sph-Mip}$ & 29.95 & 0.943 & 0.090 & 3.61
 \\
    w/o mipmap & 30.12 & 0.945 & 0.091 & 5.12
 \\
    w/o DS & 31.79 & 0.957 & 0.062 & 2.57
 \\
    w/o $\mathbf{K} \otimes \mathbf{S}$ & 33.37 & 0.966 & 0.051 & 2.38
 \\
    Ours & \textbf{34.00} & \textbf{0.969} & \textbf{0.046} & \textbf{2.21}
 \\
\end{tabular}
}
\caption{\textbf{Ablation study of our model on the synthetic datasets.}}
\label{table:Ablation studies}
\end{table}

We now perform ablation studies on the Shiny Blender~\cite{verbin2022refnerf} and NeRF Synthetic~\cite{mildenhall2020nerf} datasets. Tab.~\ref{table:Ablation studies} reports quantitative results for deferred shading, $\operatorname{Sph-Mip}$ encoding, and directional factorization. Fig.~\ref{fig:Ablation study} shows ablation comparisons for novel view synthesis and surface reconstruction.

\noindent \textbf{Sph-Mip.} We first analyze the effect of the $\operatorname{Sph-Mip}$ by directly feeding G-buffer components to the decoding MLP: $f_\Theta(\mathbf{M}, \mathbf{N}, \mathbf{d})$, where $\mathbf{d} \in \mathbb{R}^{H \times W \times 3}$ is the view directions. As shown in Fig.~\ref{fig:Ablation study} and Tab.~\ref{table:Ablation studies}, compared to directly using the G-buffer as input, our $\operatorname{Sph-Mip}$ encoding effectively models high-frequency view-dependent appearance.

\noindent \textbf{Mipmap.} To verify the effectiveness of the multi-level spherical feature grid strategies, we replace the mipmap $\mathcal{M}$ with a 2D feature map of the same shape as the base level of the mipmap $\mathcal{M}^{L_0}$ (\ie, w/o mipmap). Fig.~\ref{fig:Ablation study} shows that the method without mipmap fails to recover accurate geometry and produces artifacts when rendering rough surfaces, primarily because real-world scenes typically do not consist of a single material.

\noindent \textbf{Deferred Shading.} We ablate deferred shading (\ie, w/o DS) by applying the standard volume rendering. As shown in Fig.~\ref{fig:Ablation study}, deferred shading provides more accurate specular reflections and better surface reconstruction quality.

\noindent \textbf{Directional Factorization.} We study the proposed directional factorization (\ie, w/o $\mathbf{K} \otimes \mathbf{S}$). We directly used the directional feature $\mathbf{S}$ as input to the shader: $f_\Theta(\mathbf{S})$. As shown in Fig.~\ref{fig:Ablation study}, inter-reflections cannot be reconstructed using only the far-field feature $\mathbf{S}$.

\begin{table}[t]
    \centering
    \resizebox{0.405\textwidth}{!}{ 

    \begin{tabular}{lc|c}
    \multicolumn{1}{c}{ } & \multicolumn{1}{c|}{Rendering Speed} & \multicolumn{1}{c}{Train Time}
 \\ \hline 
    3DGS~\cite{kerbl3Dgaussians} & 1.00$\times$ & 1.00$\times$
 \\
    GaussianShader~\cite{jiang2024gaussianshader} & $0.17 \times$  & 11.05$\times$ 
 \\
    3iGS~\cite{tang20243igs} & $0.44 \times$  & 2.07$\times$ 
 \\
    3DGS-DR~\cite{ye2024gsdr} & $0.93 \times$  & 3.25$\times$ 
 \\
    Ours & $0.37 \times$  & 2.63$\times$ 
 \\
\end{tabular}
}
\caption{\textbf{Comparisons on training and rendering (test) speed.} We consider the 3DGS~\cite{kerbl3Dgaussians} as a baseline and normalise the speed. With our GPU device, 3DGS takes about 14.5 min for training, and achieves 125 FPS for rendering.
}
\label{table:speed}
\vspace{-4pt}
\end{table}

\section{Conclusion}

We have presented \textit{Ref-GS} to address view-dependent effects in 2D Gaussian Splatting, enabling photorealistic rendering and precise geometry recovery for open-world scenes. Our technical contribution is a novel deferred Gaussian rendering pipeline that integrates a spherical Mip-grid to efficiently represent surface roughness and employs a geometry-lighting factorization to explicitly connect geometry and lighting through the vector outer product.

\section*{Acknowledgements}
This work is supported by the National Natural Science Foundation of China (NSFC No.62272184) and ``Pioneer'' and ``Leading Goose'' R\&D Program of Zhejiang (No.2024C01161). The computation is completed in the HPC Platform of Huazhong University of Science and Technology.


{
    \small
    \bibliographystyle{ieeenat_fullname}
    \bibliography{main}
}

\clearpage
\setcounter{page}{1}
\maketitlesupplementary
\appendix

This supplementary material provides additional information and experiment results pertaining to the main paper including detailed descriptions of the training process, and more visual results to complement the experiments reported in the main manuscript. 

For more information regarding the method, we highly encourage readers to watch our video provided in the supplemental {\href{https://ref-gs.github.io/}{\textbf{webpage}}, where our method produces results with better specular reflection reconstruction.

\section{Implementation Details}
For training, we use the PyTorch~\cite{2019PyTorch} framework and train on a single Tesla V100 with 32GB of memory. Our code is build upon the 2D Gaussian Splatting (2DGS)~\cite{Huang2DGS2024} codebase. For real scenes, we propose using the same spherical domain strategy as 3DGS-DR~\cite{ye2024gsdr} to train our model for a fair evaluation. This approach can reduce background interference during training. Background objects, captured from only limited viewpoints, exhibit similar behavior to reflective objects, which interferes with the fitting of our $\operatorname{Sph-Mip}$.

\subsection{Network}

The goal of the shallow MLP $f_\Theta$ is is to non-linearly map the directional feature $\mathbf{S} \in \mathbb{R}^{H \times W \times 16}$ produced by the $\operatorname{Sph-Mip}$ encoding and the high-dimensional intermediate tensor  $\mathbf{K} \otimes \mathbf{S}$ has a shape of $H \times W \times 64$. Our MLP accepts an input having $16+64$ feature dimensions. The input is fed into a 2-layer MLP with 256 neurons per hidden layer in them followed by ReLU~\cite{relu} activation functions. The output is fed into a output head predicts the view-dependent radiance with a exponential function output layer. Finally, we apply gamma tone mapping~\cite{anderson1996proposal} $\gamma(\cdot)$ to convert the colors into the sRGB space before calculating the rendering loss:

\begin{equation}
    \mathbf{I} = \gamma(\mathbf{I}_d + f_\Theta(\mathbf{S}, \mathbf{K} \otimes \mathbf{S}))
\end{equation}

\begin{figure}[t]
\centering
\includegraphics[width=1.0\linewidth]{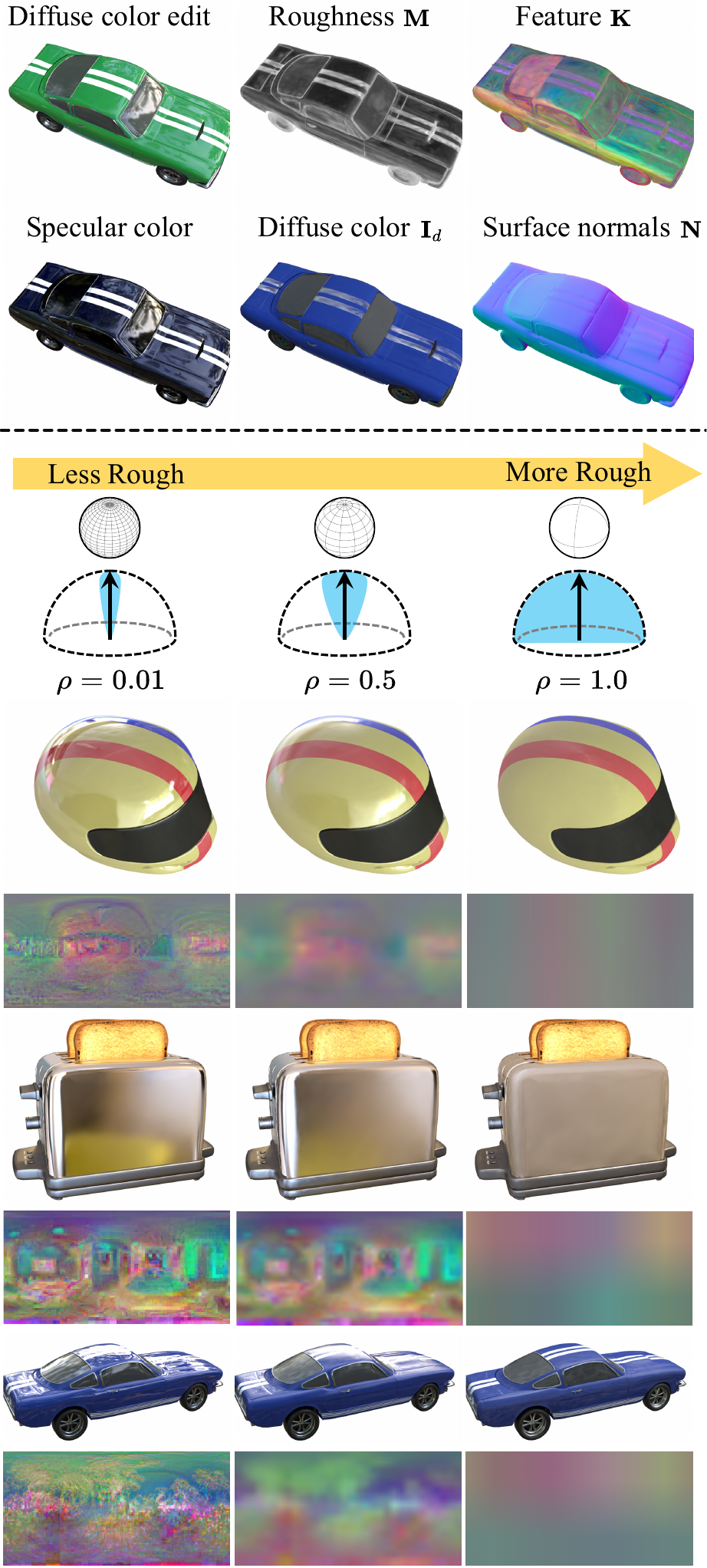}
\caption{
\textbf{Visualization of the Scene Decompositions and Material Editing}. Our model decomposes the appearance of synthetic scenes into interpretable components. \textit{Ref-GS} effectively separates view-independent diffuse colors and view-dependent specular colors from multi-view training images. Furthermore, we can edit the diffuse color of the car without affecting the specular reflections on its glossy surface (top row). By modifying roughness $\rho$, we can obtain directional feature $\mathbf{s}$ at different levels can be obtained through $\operatorname{Sph-Mip}$ interpolation (bottom row).
}
\label{fig:Editing}
\vspace{-20pt}
\end{figure}

\subsection{Optimization}

The per-Gaussian position $\mu \in \mathbb{R}^3$, scale $s \in \mathbb{R}^2$ and covariance as rotation $q \in \mathbb{R}^4$, opacity $\alpha \in \mathbb{R}$, diffuse color $\mathbf{c}_d \in \mathbb{R}^3$, roughness $\rho \in [0, 1]$, feature $\mathbf{f} \in \mathbb{R}^4$ are optimized together with the network weights for the base MLP and the output head for view-dependent radiance. We use the Adam~\cite{kingma2014adam} optimizer with default parameters. Further, we follow the default splitting and pruning schedule proposed by the original 2DGS.

\subsection{Losses}

We have multiple loss terms in our training pipeline that are mainly adapted from 2DGS that we will briefly outline them and their weighting here. As in 2DGS, we use $\mathcal{L}_1$ loss and D-SSIM~\cite{Wang_Bovik_Sheikh_Simoncelli_2004} loss for supervising RGB color, with $\lambda = 0.2$:
\begin{equation}
\mathcal{L}_\text{rgb} = (1-\lambda) \mathcal{L}_1 + \lambda \mathcal{L}_\text{D-SSIM}.
\label{eq:RGBLoss}
\end{equation}
Following 2DGS, depth distortion loss and normal consistency loss are adopted to refine the geometry property of the 2DGS representation of the scene. 
\begin{equation}
\mathcal{L}_\text{d} = \sum_{i,j}\omega_i\omega_j|z_i-z_j| \hspace{0.5cm} \mathcal{L}_{n} = \sum_{i} \omega_i (1-\mathbf{n}_i^\top\mathbf{\widehat{N}})
\end{equation}
Here, $\omega_i$ represents the blending weight of the $i^{th}$ intersection. $z_i$ denotes the depth of the intersection points.
$\mathbf{n}_i$ is the normal of the splat facing the camera. $\mathbf{\widehat{N}}$ is the normal estimated by the gradient of the depth map. The total loss is given as:
\begin{equation}
    \mathcal{L} = \mathcal{L}_\text{rgb} + \lambda_{d} \mathcal{L}_\text{d} + \lambda_{n} \mathcal{L}_\text{n}
\end{equation}
We empirically set $\lambda_{d} = 100$, $\lambda_{n} = 0.05$.

\section{Limitations}
While our approach demonstrates effective performance with a lightweight MLP for final color prediction, it results in slower rendering speeds compared to 2DGS and is challenging to integrate into standard CG rendering engines due to its reliance on a neural decoder. However, conversion techniques like textured mesh baking can facilitate integration and benefit from our reconstruction pipeline's thin surface modeling and rendering capabilities.

\section{Additional Results}
In this section, we present additional visual results to demonstrate the capability of \textit{Ref-GS} in reconstructing and rendering glossy surfaces, showcasing superior visual quality and accurate predicted normals for specular reflections across diverse scenes in the proposed dataset.

\begin{table}[h]
    \centering
    \resizebox{0.49\textwidth}{!}{ 
    \begin{tabular}{lcccccccc}
    \hline

    & \multicolumn{7}{c}{Shiny Blender}\\ 
    & Car & Ball & Helmet & Teapot & Toaster & Coffee & Avg.\\ \hline    
    & \multicolumn{7}{c}{MAE$^\circ$$\downarrow$} \\ \hline

    NVDiffRec~\cite{nvdiffrec_2022_CVPR} & 11.78 & 32.67 & 21.19 & 5.55 & 16.04 & 15.05 & 17.05 
 \\
    Ref-NeRF~\cite{verbin2022refnerf} & 14.93 & 1.55 & 29.48 & 9.23 & 42.87 & 12.24 & 18.38 
 \\
    ENVIDR~\cite{liang2023envidr} & 7.10\cellcolor{yellow!25} & 0.74\cellcolor{red!25} & 1.66\cellcolor{red!25} & 2.47\cellcolor{yellow!25} & 6.45\cellcolor{orange!25} & 9.23 & 4.61\cellcolor{yellow!25}
 \\
    GaussianShader~\cite{jiang2024gaussianshader} & 14.05 & 7.03 & 9.33 & 7.17 & 13.08 & 14.93 & 10.93 
 \\
    GS-IR~\cite{liang2023gsir} & 28.31 & 25.79 & 25.58 & 15.35 & 33.51 & 15.38 & 23.99 
 \\
    RelightGS~\cite{R3DG2023} & 26.02 & 22.44 & 19.63 & 9.21 & 28.17 & 13.39 & 19.81 
 \\
    3iGS~\cite{tang20243igs} & 11.79 & 31.78 & 16.72 & 2.61 & 21.12 & 8.80\cellcolor{yellow!25} & 15.47 
 \\
    3DGS-DR~\cite{ye2024gsdr} & 2.32\cellcolor{orange!25} & 0.85\cellcolor{orange!25} & 1.67\cellcolor{orange!25} & 0.53\cellcolor{red!25} & 6.99\cellcolor{yellow!25} & 2.21\cellcolor{red!25} & 2.43\cellcolor{orange!25} 
 \\
    GS-ROR~\cite{zhu2024gs} & 11.98 & 0.92\cellcolor{yellow!25} & 4.10 & 5.88 & 8.24 & 12.24 & 7.23
 \\ \hline
    Ours & 2.02\cellcolor{red!25} & 1.05 & 1.99\cellcolor{yellow!25} & 0.69\cellcolor{orange!25} & 3.92\cellcolor{red!25} & 3.61\cellcolor{orange!25} & 2.21\cellcolor{red!25} 
 \\ \hline \hline
\end{tabular}
}
\caption{Quantitative Mean Angular Error in degrees (MAE$^\circ$$\downarrow$) of individual scenes on Shiny Blender~\cite{verbin2022refnerf} dataset. \colorbox{red!25}{Red}, \colorbox{orange!25}{Orange}, and \colorbox{yellow!25}{Yellow} indicate the first, second, and third best performing methods for each scene.}
\label{table:normal}
\end{table}

\subsection{Shiny Blender Dataset}
Tab.~\ref{table:normal} provides the results on normal estimation for all scenes on Shiny Blender~\cite{verbin2022refnerf} dataset. For 3iGS~\cite{tang20243igs}, we use grad normals derived from the rendered depth map for evaluation.

\subsection{Glossy Synthetic Dataset}
We present the novel view synthesis results on the Glossy Synthetic~\cite{liu2023nero} dataset. The quantitative evaluation in terms of Peak Signal-to-Noise Ratio (PSNR), Structural Similarity Index Measure (SSIM)~\cite{Wang_Bovik_Sheikh_Simoncelli_2004}, and Learned Perceptual Image Patch Similarity (LPIPS)~\cite{Zhang_Isola_Efros_Shechtman_Wang_2018}. is present in Tab.~\ref{tab:Glossy Synthetic}. Our approach outperforms the existing Gaussian-based methods~\cite{ye2024gsdr, zhu2024gs, tang20243igs, jiang2024gaussianshader} on most scenes.

\begin{table}[h]
    \centering
    \resizebox{0.49\textwidth}{!}{ 

    \begin{tabular}{lccccccc}
    \hline
    & \multicolumn{7}{c}{Glossy Synthetic} \\
    & Bell & Cat & Luyu & Potion & Tbell & Teapot & Avg.\\ \hline 
             
    & \multicolumn{7}{c}{PSNR$\uparrow$} \\ \hline
    Ref-NeRF~\cite{verbin2022refnerf} & 30.02 & 29.76 & 25.42 & 30.11 & 26.91 & 22.77 & 27.50 
 \\
    NeRO~\cite{liu2023nero} & — & — & — & — & — & — & —
 \\
    ENVIDR~\cite{liang2023envidr} & 30.88 & 31.04 & 28.03 & 32.11\cellcolor{orange!25} & 28.64\cellcolor{yellow!25} & 26.77\cellcolor{red!25} & 29.58 
 \\
    3DGS~\cite{kerbl3Dgaussians} & 25.11 & 31.36 & 26.97 & 30.16 & 23.88 & 21.51 & 26.50 
 \\
    GaussianShader~\cite{jiang2024gaussianshader} & 28.07 & 31.81\cellcolor{yellow!25} & 27.18 & 30.09 & 24.48 & 23.58 & 27.54
 \\
    3iGS~\cite{tang20243igs} & 25.60 & 30.93 & 27.17 & 29.50 & 23.94 & 21.17 & 26.39 
 \\
    3DGS-DR~\cite{ye2024gsdr} & 31.84\cellcolor{red!25} & 33.39\cellcolor{red!25} & 28.62\cellcolor{orange!25} & 31.74\cellcolor{yellow!25} & 27.65 & 25.44 & 29.78\cellcolor{orange!25} 
 \\
    GS-ROR~\cite{zhu2024gs} & 31.53\cellcolor{yellow!25} & 31.72 & 28.53\cellcolor{yellow!25} & 30.51 & 29.48\cellcolor{orange!25} & 26.41\cellcolor{yellow!25} & 29.70\cellcolor{yellow!25} 
 \\ \hline
    Ours & 31.70\cellcolor{orange!25} & 33.15\cellcolor{orange!25} & 29.46\cellcolor{red!25} & 32.64\cellcolor{red!25} & 30.08\cellcolor{red!25} & 26.47\cellcolor{orange!25} & 30.59\cellcolor{red!25} 
 \\ \hline \hline
 
    & \multicolumn{7}{c}{SSIM$\uparrow$} \\ \hline

    Ref-NeRF~\cite{verbin2022refnerf} & 0.941 & 0.944 & 0.901 & 0.933 & 0.947 & 0.897 & 0.927 
 \\
    NeRO~\cite{liu2023nero} & 0.965\cellcolor{orange!25} & 0.962 & 0.914 & 0.950\cellcolor{yellow!25} & 0.968\cellcolor{red!25} & 0.977\cellcolor{red!25} & 0.956\cellcolor{orange!25} 
 \\
    ENVIDR~\cite{liang2023envidr} & 0.954 & 0.965 & 0.931\cellcolor{yellow!25} & 0.960\cellcolor{red!25} & 0.947 & 0.957\cellcolor{orange!25} & 0.952 
 \\
    3DGS~\cite{kerbl3Dgaussians} & 0.908 & 0.959 & 0.916 & 0.938 & 0.900 & 0.881 & 0.917 
 \\
    GaussianShader~\cite{jiang2024gaussianshader} & 0.919 & 0.961 & 0.914 & 0.936 & 0.898 & 0.901 & 0.922 
 \\
    3iGS~\cite{tang20243igs} & 0.898 & 0.960 & 0.916 & 0.936 & 0.896 & 0.869 & 0.913 
 \\
    3DGS-DR~\cite{ye2024gsdr} & 0.964\cellcolor{yellow!25} & 0.976\cellcolor{red!25} & 0.938\cellcolor{orange!25} & 0.957\cellcolor{orange!25} & 0.948 & 0.939 & 0.954\cellcolor{yellow!25} 
 \\
    GS-ROR~\cite{zhu2024gs} & 0.969\cellcolor{red!25} & 0.967\cellcolor{yellow!25} & 0.938\cellcolor{orange!25} & 0.950\cellcolor{yellow!25} & 0.965\cellcolor{orange!25} & 0.947\cellcolor{yellow!25} & 0.956\cellcolor{orange!25} 
 \\ \hline
    Ours & 0.965\cellcolor{orange!25} & 0.973\cellcolor{orange!25} & 0.946\cellcolor{red!25} & 0.957\cellcolor{orange!25} & 0.956\cellcolor{yellow!25} & 0.944 & 0.957\cellcolor{red!25} 
 \\ \hline \hline

    & \multicolumn{7}{c}{LPIPS$\downarrow$} \\ \hline
    Ref-NeRF~\cite{verbin2022refnerf} & 0.102 & 0.104 & 0.098 & 0.084 & 0.114 & 0.098 & 0.100
 \\
    NeRO~\cite{liu2023nero} & 0.056 & 0.052 & 0.072 & 0.084 & 0.046 & 0.028\cellcolor{red!25} & 0.056\cellcolor{red!25} 
 \\
    ENVIDR~\cite{liang2023envidr} & 0.054\cellcolor{yellow!25} & 0.049\cellcolor{yellow!25} & 0.059\cellcolor{yellow!25} & 0.072\cellcolor{red!25} & 0.069\cellcolor{red!25} & 0.041\cellcolor{orange!25} & 0.057\cellcolor{orange!25} 
 \\
    3DGS~\cite{kerbl3Dgaussians} & 0.104 & 0.062 & 0.064 & 0.093 & 0.125 & 0.102 & 0.092 
 \\
    GaussianShader~\cite{jiang2024gaussianshader} & 0.098 & 0.056 & 0.064 & 0.088 & 0.122 & 0.091 & 0.087
 \\
    3iGS~\cite{tang20243igs} & 0.104 & 0.057 & 0.064 & 0.089 & 0.119 & 0.103 & 0.089 
 \\
    3DGS-DR~\cite{ye2024gsdr} & 0.044\cellcolor{red!25} & 0.039\cellcolor{red!25} & 0.052\cellcolor{orange!25} & 0.073\cellcolor{orange!25} & 0.070\cellcolor{orange!25} & 0.062\cellcolor{yellow!25} & 0.057\cellcolor{orange!25} 
 \\
    GS-ROR~\cite{zhu2024gs} & — & — & — & — & — & — & —
 \\ \hline
    Ours & 0.049\cellcolor{orange!25} & 0.041\cellcolor{orange!25} & 0.046\cellcolor{red!25} & 0.076\cellcolor{yellow!25} & 0.073\cellcolor{yellow!25} & 0.064 & 0.058\cellcolor{yellow!25} 
 \\ \hline \hline
    \end{tabular}
}
\caption{Quantitative results of individual scenes on Glossy Synthetic~\cite{liu2023nero} dataset. \colorbox{red!25}{Red}, \colorbox{orange!25}{Orange}, and \colorbox{yellow!25}{Yellow} indicate the first, second, and third best performing methods for each scene.}
\label{tab:Glossy Synthetic}
\end{table}

\subsection{Glossy Real Dataset}
We present the geometry reconstruction results on the Glossy Real~\cite{liu2023nero} dataset to further validate the robustness and accuracy of our approach. We visualized the reconstruction results as shown in Fig.~\ref{fig:glossyreal}.

For a more comprehensive view of our method’s performance, please refer to the videos provided on the supplemental {\href{https://ref-gs.github.io/}{\textbf{webpage}}.

\begin{figure}[h]
\centering
\includegraphics[width=1.0\linewidth]{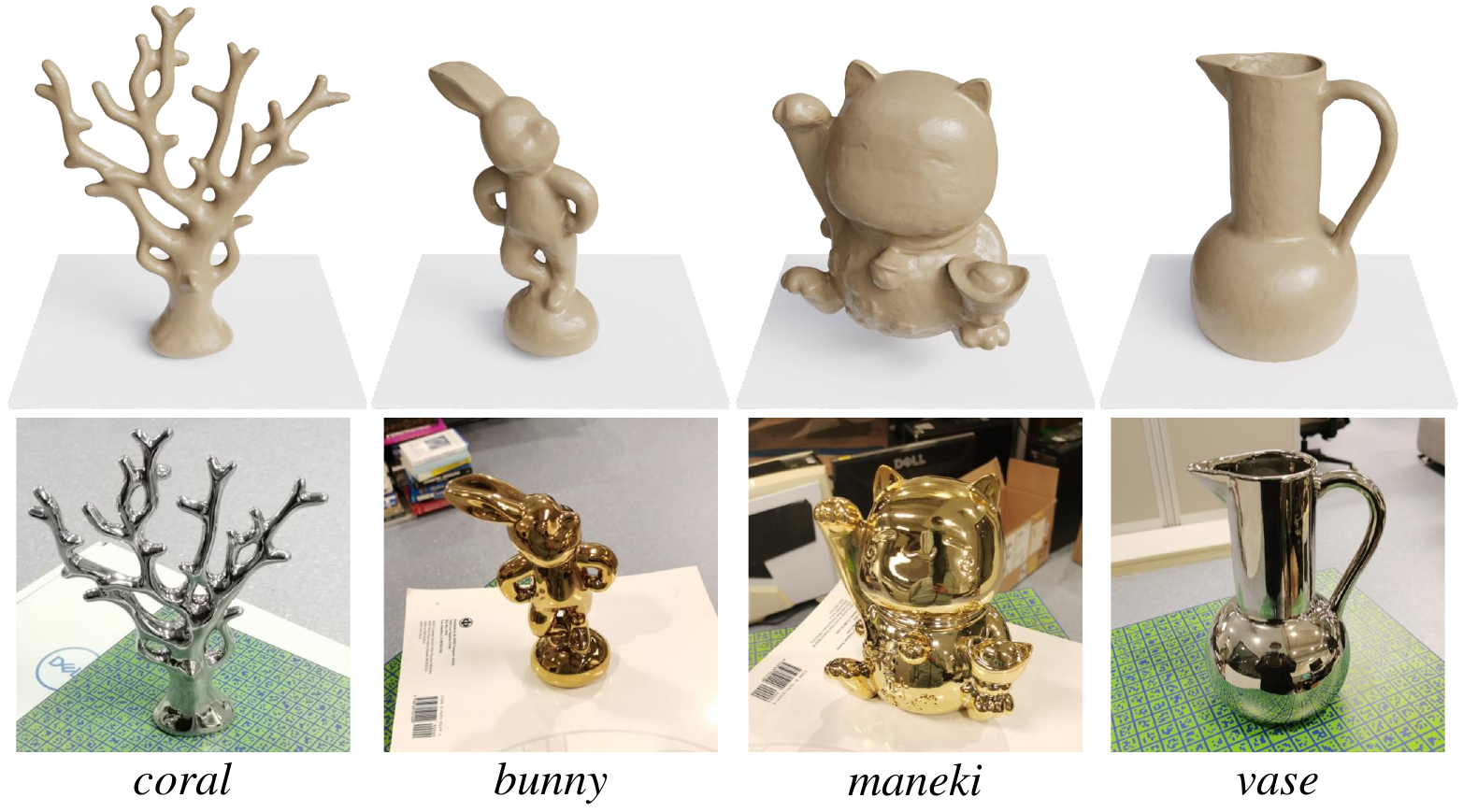}
\caption{Images, ground-truth and reconstructed surfaces of the Glossy Real~\cite{liu2023nero} dataset.
}
\label{fig:glossyreal}
\end{figure}

\subsection{NeRF Synthetic Dataset}
Quantitative results on the NeRF Synthetic~\cite{mildenhall2020nerf} dataset are reported in Tab.~\ref{table:nerf}. Our approach achieves numerically and visually comparable results with Gaussian-based methods~\cite{ye2024gsdr, zhu2024gs, tang20243igs, jiang2024gaussianshader}, demonstrating the effectiveness of our method in rendering general objects.

\begin{table}[h]
    \centering
    \resizebox{0.5\textwidth}{!}{ 

    \begin{tabular}{lccccccccc}
    \hline
    & \multicolumn{9}{c}{NeRF Synthetic} \\
    & Chair & Drums & Lego & Mic & Materials & Ship & Hotdog & Ficus & Avg.\\ \hline 
             
    & \multicolumn{9}{c}{PSNR$\uparrow$} \\ \hline
    NeRF~\cite{mildenhall2020nerf} & 33.00 & 25.01 & 32.54 & 32.91 & 29.62 & 28.65 & 36.18 & 30.13 & 31.01 
 \\
    Ref-NeRF~\cite{verbin2022refnerf} & 33.98 & 25.43 & 35.10 & 33.65 & 27.10 & 29.24 & 37.04 & 28.74 & 31.29 
 \\
    VolSDF~\cite{yariv2021volume} & 30.57 & 20.43 & 29.46 & 30.53 & 29.13 & 25.51 & 35.11 & 22.91 & 27.96 
 \\
    ENVIDR~\cite{liang2023envidr} & 31.22 & 22.99 & 29.55 & 32.17 & 29.52 & 21.57 & 31.44 & 26.60 & 28.13 
 \\
    3DGS~\cite{kerbl3Dgaussians} & 35.82\cellcolor{red!25} & 26.17\cellcolor{yellow!25} & 35.69\cellcolor{yellow!25} & 35.34\cellcolor{yellow!25} & 30.00\cellcolor{orange!25} & 30.87\cellcolor{orange!25} & 37.67\cellcolor{orange!25} & 34.83\cellcolor{orange!25} & 33.30\cellcolor{orange!25} 
 \\
    GaussianShader~\cite{jiang2024gaussianshader} & 33.70 & 25.50 & 32.99 & 34.07 & 28.87 & 28.37 & 35.29 & 33.05 & 31.48 
 \\
    3iGS~\cite{tang20243igs} & 35.59\cellcolor{yellow!25} & 26.75\cellcolor{red!25} & 35.94\cellcolor{orange!25} & 36.01\cellcolor{red!25} & 30.00\cellcolor{orange!25} & 31.12\cellcolor{red!25} & 37.98\cellcolor{red!25} & 35.40\cellcolor{red!25} & 33.60\cellcolor{red!25} 
 \\
    3DGS-DR~\cite{ye2024gsdr} & 35.60\cellcolor{orange!25} & 25.31 & 32.94 & 31.97 & 29.65\cellcolor{yellow!25} & 29.07 & 35.58 & 28.03 & 31.02 
 \\ \hline
    Ours & 34.66 & 26.33\cellcolor{orange!25} & 36.26\cellcolor{red!25} & 35.76\cellcolor{orange!25} & 30.99\cellcolor{red!25} & 29.67\cellcolor{yellow!25} & 37.39\cellcolor{yellow!25} & 34.52\cellcolor{yellow!25} & 33.20\cellcolor{yellow!25} 
 \\ \hline \hline

    & \multicolumn{9}{c}{SSIM$\uparrow$} \\ \hline
    NeRF~\cite{mildenhall2020nerf} & 0.967 & 0.925 & 0.961 & 0.980 & 0.949 & 0.856 & 0.974 & 0.964 & 0.947
 \\
    Ref-NeRF~\cite{verbin2022refnerf} & 0.974 & 0.929 & 0.975 & 0.983 & 0.921 & 0.864 & 0.979 & 0.954 & 0.947 
 \\
    VolSDF~\cite{yariv2021volume} & 0.949 & 0.893 & 0.951 & 0.969 & 0.954 & 0.842 & 0.972 & 0.929 & 0.932 
 \\
    ENVIDR~\cite{liang2023envidr} & 0.976 & 0.930 & 0.961 & 0.984 & 0.968\cellcolor{red!25} & 0.855 & 0.963 & 0.987\cellcolor{orange!25} & 0.953 
 \\
    3DGS~\cite{kerbl3Dgaussians} & 0.987\cellcolor{red!25} & 0.954\cellcolor{orange!25} & 0.983\cellcolor{red!25} & 0.991\cellcolor{orange!25} & 0.960 & 0.907\cellcolor{orange!25} & 0.985\cellcolor{orange!25} & 0.987\cellcolor{orange!25} & 0.969\cellcolor{orange!25} 
 \\
    GaussianShader~\cite{jiang2024gaussianshader} & 0.980 & 0.945 & 0.972 & 0.989\cellcolor{yellow!25} & 0.951 & 0.881 & 0.980 & 0.982\cellcolor{yellow!25} & 0.960 
 \\
    3iGS~\cite{tang20243igs} & 0.987\cellcolor{red!25} & 0.955\cellcolor{red!25} & 0.983\cellcolor{red!25} & 0.992\cellcolor{red!25} & 0.961\cellcolor{yellow!25} & 0.908\cellcolor{red!25} & 0.986\cellcolor{red!25} & 0.989\cellcolor{red!25} & 0.970\cellcolor{red!25} 
 \\
    3DGS-DR~\cite{ye2024gsdr} & 0.986\cellcolor{orange!25} & 0.946 & 0.978\cellcolor{yellow!25} & 0.987 & 0.958 & 0.894\cellcolor{yellow!25} & 0.982\cellcolor{yellow!25} & 0.963 & 0.962 
 \\ \hline
    Ours & 0.985\cellcolor{yellow!25} & 0.952\cellcolor{yellow!25} & 0.982\cellcolor{orange!25} & 0.991\cellcolor{orange!25} & 0.964\cellcolor{orange!25} & 0.890 & 0.984\cellcolor{yellow!25} & 0.982 & 0.966\cellcolor{yellow!25} 
 \\ \hline \hline

    & \multicolumn{9}{c}{LPIPS$\downarrow$} \\ \hline
    NeRF~\cite{mildenhall2020nerf} & 0.046 & 0.091 & 0.050 & 0.028 & 0.063 & 0.206 & 0.121 & 0.044 & 0.081 
 \\
    Ref-NeRF~\cite{verbin2022refnerf} & 0.029 & 0.073 & 0.025\cellcolor{yellow!25} & 0.018 & 0.078 & 0.158 & 0.028 & 0.056 & 0.058 
 \\
    VolSDF~\cite{yariv2021volume} & 0.056 & 0.119 & 0.054 & 0.191 & 0.048 & 0.191 & 0.043 & 0.068 & 0.096 
 \\
    ENVIDR~\cite{liang2023envidr} & 0.031 & 0.080 & 0.054 & 0.021 & 0.045 & 0.228 & 0.072 & 0.010\cellcolor{red!25} & 0.068 
 \\
    3DGS~\cite{kerbl3Dgaussians} & 0.012\cellcolor{red!25} & 0.037\cellcolor{orange!25} & 0.016\cellcolor{orange!25} & 0.006\cellcolor{orange!25} & 0.034\cellcolor{red!25} & 0.106\cellcolor{orange!25} & 0.020\cellcolor{orange!25} & 0.012\cellcolor{orange!25} & 0.030\cellcolor{orange!25} 
 \\
    GaussianShader~\cite{jiang2024gaussianshader} & 0.019 & 0.045 & 0.026 & 0.009\cellcolor{yellow!25} & 0.046 & 0.148 & 0.029 & 0.017\cellcolor{yellow!25} & 0.042 
 \\
    3iGS~\cite{tang20243igs} & 0.012\cellcolor{red!25} & 0.036\cellcolor{red!25} & 0.015\cellcolor{red!25} & 0.005\cellcolor{red!25} & 0.034\cellcolor{red!25} & 0.102\cellcolor{red!25} & 0.019\cellcolor{red!25} & 0.010\cellcolor{red!25} & 0.029\cellcolor{red!25} 
 \\
    3DGS-DR~\cite{ye2024gsdr} & 0.014\cellcolor{yellow!25} & 0.055 & 0.026 & 0.028 & 0.038\cellcolor{orange!25} & 0.129 & 0.033 & 0.055 & 0.047 
 \\ \hline
    Ours & 0.013\cellcolor{orange!25} & 0.044\cellcolor{yellow!25} & 0.016\cellcolor{orange!25} & 0.009\cellcolor{yellow!25} & 0.042\cellcolor{yellow!25} & 0.127\cellcolor{yellow!25} & 0.021\cellcolor{yellow!25} & 0.017\cellcolor{yellow!25} & 0.036\cellcolor{yellow!25} 
 \\ \hline \hline
    \end{tabular}
}
\caption{Quantitative results of individual scenes on NeRF Synthetic~\cite{mildenhall2020nerf} dataset. \colorbox{red!25}{Red}, \colorbox{orange!25}{Orange}, and \colorbox{yellow!25}{Yellow} indicate the first, second, and third best performing methods for each scene.}
\label{table:nerf}
\end{table}

\begin{figure*}[t]
\centering
\includegraphics[width=1.0\linewidth]{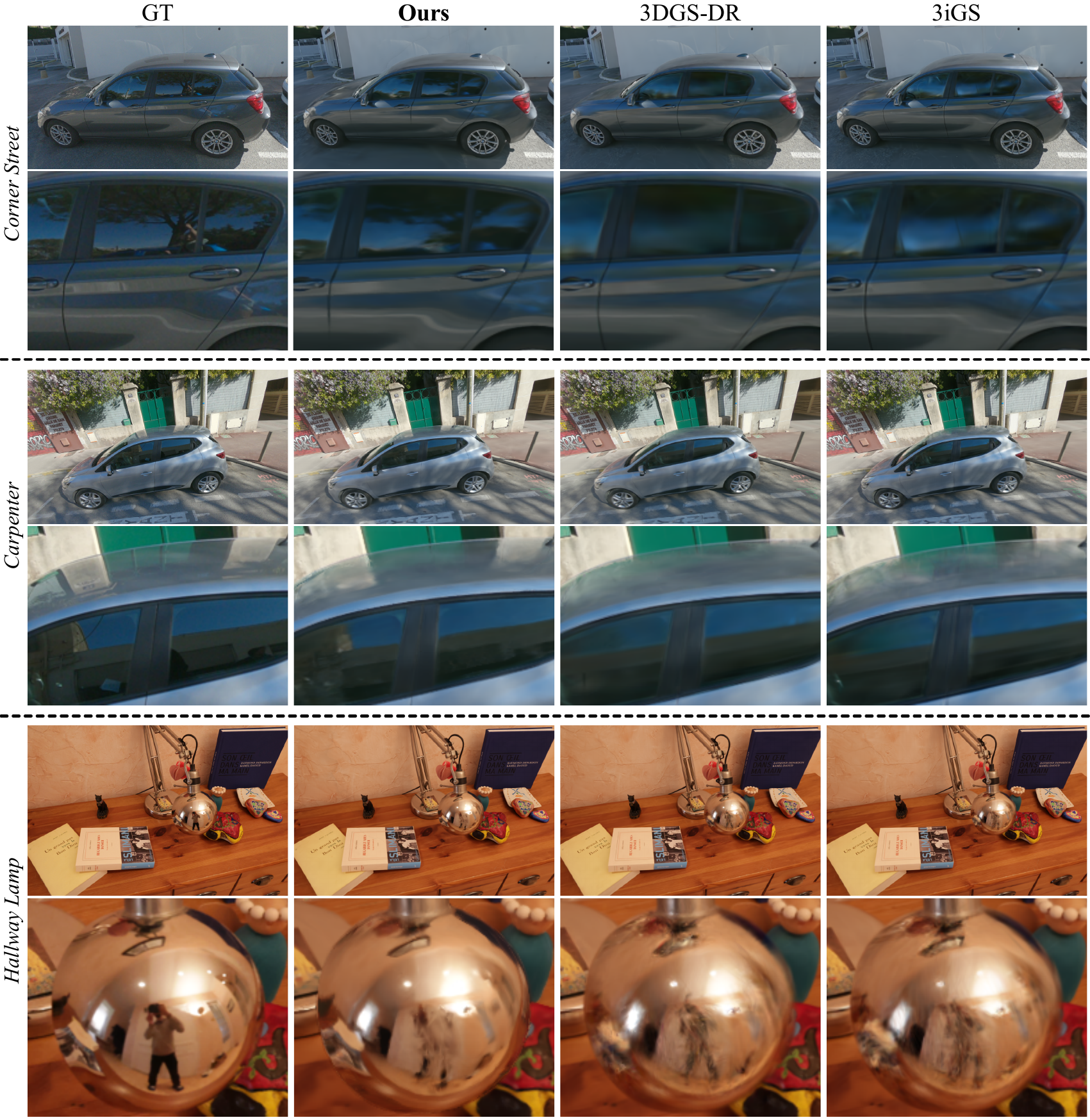}
\caption{
Additional results for intermediate component visualizations of our approach compared to 3DGS-DR~\cite{ye2024gsdr} and 3iGS~\cite{tang20243igs} on the Rodriguez \etal ~\cite{RPHD20} and Kopanas \etal ~\cite{kopanas2022neural} datasets; zoom in to see the difference. (\textit{Corner Street}, \textbf{1st row}) Our approach effectively simulates realistic reflections on the car body and windshield. (\textit{Carpenter}, \textbf{2nd row}) Reflections of distant scenes on the car roof are rendered with impressive accuracy. (\textit{Hallway Lamp}, \textbf{3rd row}) High-frequency details are well-preserved, enabling the realistic depiction of near-field content, including precise reflections.
}
\label{fig: real}
\end{figure*}

\subsection{Additional Ablation Results}
We provide more ablation results of on synthesized test in Tab.~\ref{tb:moreablations}. To more clearly demonstrate the distinct advantages of the 2D Gaussian representation, we replaced 2DGS~\cite{Huang2DGS2024} with 3DGS~\cite{kerbl3Dgaussians}, using the shortest axis as the plane normal while keeping the rest unchanged for comparison, as shown in the first two rows of Tab.~\ref{tb:moreablations}. Furthermore, We have conducted ablation studies on the grid size $N$ of $\operatorname{Sph-Mip}$, as shown in Tab.~\ref{tb:moreablations}. Notably, 3DGS-DR\cite{ye2024gsdr} improves the performance of GaussianShader\cite{jiang2024gaussianshader} by introducing deferred shading with a simple shading model. ``w/o $\mathbf{K} \otimes \mathbf{S}$'' demonstrates that the $\operatorname{Sph-Mip}$ encoding can further enhance rendering quality. Additionally, the results of ``w/o DS'' demonstrate that our method outperforms the explicit BRDF of GaussianShader.

\begin{table}[h]
\centering
\resizebox{0.5\textwidth}{!}{%
\begin{tabular}{l|cccccccc}
\hline
\textbf{} & Chair & Drums & Lego & Mic & Materials & Ship & Hotdog & Ficus \\
\hline
\textbf{Ours} & \cellcolor{red!25}34.66 & \cellcolor{red!25}26.33 & \cellcolor{red!25}36.26 & \cellcolor{red!25}35.76 & \cellcolor{orange!25}30.99 & \cellcolor{red!25}29.67 & \cellcolor{red!25}37.39 & \cellcolor{red!25}34.52 \\
w/ 3DGS & \cellcolor{orange!25}34.15 & \cellcolor{orange!25}25.86 & \cellcolor{yellow!25}34.74 & \cellcolor{yellow!25}34.73 & \cellcolor{red!25}31.32 & \cellcolor{orange!25}29.52 & \cellcolor{orange!25}36.78 & \cellcolor{orange!25}33.10\\
\hline

$\operatorname{Sph-Mip}$ $N$=8 & \cellcolor{blue!25}34.67 & \cellcolor{blue!10}26.34 & \cellcolor{blue!25}35.83 & \cellcolor{blue!25}35.23 & \cellcolor{blue!10}30.91 & \cellcolor{blue!25}29.26 & \cellcolor{blue!25}37.19 & \cellcolor{blue!25}34.11 \\

$\operatorname{Sph-Mip}$ $N$=7 & \cellcolor{blue!0}34.64 & \cellcolor{blue!25}26.36 & \cellcolor{blue!10}35.80 & \cellcolor{blue!10}35.17 & \cellcolor{blue!25}31.00 & \cellcolor{blue!10}29.23 & \cellcolor{blue!0}37.10 & \cellcolor{blue!10}34.10 \\

$\operatorname{Sph-Mip}$ $N$=6 & \cellcolor{blue!10}35.65 & \cellcolor{blue!0}26.17 & \cellcolor{blue!0}35.74 & \cellcolor{blue!0}35.04 & \cellcolor{blue!0}30.39 & \cellcolor{blue!0}29.16 & \cellcolor{blue!10}37.13 & \cellcolor{blue!10}34.10 \\
\hline
w/o DS  & 33.75 & \cellcolor{yellow!25}25.85 & 33.99 & \cellcolor{orange!25}35.16 & 29.25 & 28.89 & 36.11 & 32.15\\ 
w/o $\mathbf{K} \otimes \mathbf{S}$ & \cellcolor{yellow!25}34.08 & 25.71 & \cellcolor{orange!25}35.19 & 34.21 & \cellcolor{yellow!25}29.77 & \cellcolor{yellow!25}29.10 & \cellcolor{yellow!25}36.62 & \cellcolor{yellow!25}32.47\\

\hline
\end{tabular}%
}
\caption{Per-scene PSNR comparison on NeRF Synthetic dataset. w/ 3DGS: Using 3DGS as the representation of our \textit{Ref-GS} with the rest unchanged.}
\label{tb:moreablations}
\vspace{-8pt}
\end{table}

\subsection{Additional Results on Real-World Captures}
In this section, we extend the evaluation of our proposed method to include its performance on Rodriguez \etal ~\cite{RPHD20} and Kopanas \etal ~\cite{kopanas2022neural} datasets. The qualitative comparison in Fig.~\ref{fig: real} shows that \textit{Ref-GS} extends well to real scenes, producing clearer specular reflections of the complex real-world environments compared to the existing Gaussian-based methods.

\subsection{Scene Decompositions and Editing}
Fig.~\ref{fig:Editing} illustrates the rendering decomposition results of the scene. For reflective objects exhibiting strong specular effects, our approach can effectively decompose both the view-independent diffuse color and view-dependent specular color. Furthermore, the predicted material properties (\textit{e.g.}, roughness $\rho$) and far-field lighting $\mathcal{M}$ are also very reasonable. Additionally, we can plausibly modify the roughness of the scenes by adjusting the $\rho$ values.

\subsection{Supplementary Video Results}
For a more comprehensive understanding of the performance of our approach, please refer to the supplementary videos provided. Additionally, we have created an interactive {\href{https://ref-gs.github.io/}{\textbf{webpage}} to vividly showcase the capabilities of our approach.

\end{document}